\documentclass[letterpaper, 10 pt]{IEEEtran}
\IEEEoverridecommandlockouts

\usepackage{xurl}
\usepackage[x11names, table]{xcolor}
\usepackage[colorlinks, bookmarksopen, bookmarksnumbered, citecolor=Red4, urlcolor=Turquoise4, linkcolor=Green4]{hyperref}

\usepackage{amsmath}
\usepackage{amssymb}
\usepackage{graphicx}
\usepackage{booktabs}
\usepackage{multirow}
\usepackage{dsfont}
\usepackage{mathtools}
\usepackage[table-alignment-mode=format]{siunitx}
\usepackage[caption=false]{subfig}
\usepackage{balance}
\usepackage[ruled, vlined]{algorithm2e}

\usepackage{natbib}

\title{\Large \bf 
Autonomous loading of ore piles with Load-Haul-Dump machines using Deep Reinforcement Learning}

\author{Rodrigo Salas$^{1}$, Francisco Leiva$^{1}$, and Javier Ruiz-del-Solar$^{1}$%
\thanks{This work was supported by FONDECYT project 1201170, and ANID-PIA project AFB230001.}%
\thanks{$^1$Advanced Mining Technology Center (AMTC) and Department of Electrical Engineering, Universidad de Chile, Tupper 2007, Santiago, Chile.}%
\thanks{\tt\small{\{rodrigo.salas, francisco.leiva, jruizd\}@ing.uchile.cl}}%
}

\begin{document}

\maketitle

\begin{abstract}
This work presents a deep reinforcement learning-based approach to train controllers for the autonomous loading of ore piles with a Load-Haul-Dump (LHD) machine. These controllers must perform a complete loading maneuver, filling the LHD's bucket with material while avoiding wheel drift, dumping material, or getting stuck in the pile. The training process is conducted entirely in simulation, using a simple environment that leverages the Fundamental Equation of Earth-Moving Mechanics so as to achieve a low computational cost. Two different types of policies are trained: one with a hybrid action space and another with a continuous action space. The RL-based policies are evaluated both in simulation and in the real world using a scaled LHD and a scaled muck pile, and their performance is compared to that of a heuristics-based controller and human teleoperation. Additional real-world experiments are performed to assess the robustness of the RL-based policies to measurement errors in the characterization of the piles. Overall, the RL-based controllers show good performance in the real world, achieving fill factors between 71-94\%, and less wheel drift than the other baselines during the loading maneuvers. A video showing the training environment and the learned behavior in simulation, as well as some of the performed experiments in the real world, can be found in \url{https://youtu.be/jOpA1rkwhDY}.
\end{abstract}


\section{Introduction}
\label{sec:introduction}

Automating the operation of mining equipment has the potential to improve both the productivity of mines and the safety of their workers~\citep{ali2020artificial}. As we move toward deep underground mining~\citep{ghorbani2023moving}, workers are more exposed than before to risks such as rock falls, and continued exposure to dust and combustion gases \citep{salvador2020automation, xiao2022research}. In underground mining, the most critical earth-moving vehicle is the Load-Haul-Dump machine (LHD, and also known as scoop tram), which is responsible for loading and transporting minerals at the production level~\citep{tatiya2005surface}.

An LHD is an articulated vehicle with a front bucket designed to load, haul and dump material extracted from ore/muck piles, placed in so-called ``draw points''. The operating cycle of an LHD, also referred to as the ``\textit{V-cycle}''~\citep{filla2014study} or ``\textit{short loading cycle}''~\citep{dadhich2016key}, has three main steps: (i) loading or excavating material, (ii) navigating, and (iii) dumping the material. Both autonomous navigation and autonomous excavation have been studied with varying degrees of success, with autonomous navigation systems already being used in underground operations worldwide~\citep{rct_automation, sandvik_automation, epiroc_automation}. Autonomous excavation, on the other hand, is still in its early stages of development~\citep{xiao2022research}. 

\begin{figure}
    \centering
    \includegraphics[width=\linewidth]{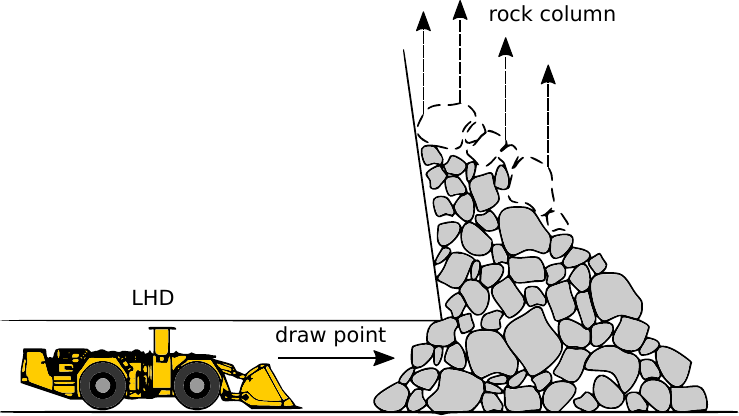}
    \caption{Diagram of the side view of a sublevel stoping draw point from which an LHD has to load material (adapted from \cite{tampier2021autonomous}, with permission of the authors).}
    \label{fig:tampier_muck_pile}
\end{figure}

The delay in progress in autonomous loading (as compared to autonomous navigation) may be due to the challenging nature of the task: a system for autonomous excavation must deal with the unpredictable nature of the interaction between the LHD and the material~\citep{luengo1998modeling}, and must be able to account for different mine layouts (e.g., with varying sizes of both tunnels and draw points) and with progressive machine wear (e.g. in the LHD's tires, because of drifting). Loading from draw points, such as those used in block/panel caving and sublevel stoping mines (see Fig.~\ref{fig:tampier_muck_pile}), is a complex process, not only because the interaction between the bucket of the LHD and the material is difficult to model, but also because the column of blasted rock exerts large compressive forces on the open face of the draw point~\citep{tampier2021autonomous}.

Autonomous excavation with front-loading earth-moving vehicles (e.g., LHDs, front loaders, or skid-steer loaders) has been addressed following several different approaches~\citep{dadhich2016key}, e.g., utilizing trajectory planners for the bucket inside the muck pile~\citep{meng2019bucket, chen2022research}, force and compliance controllers~\citep{dobson2017admittance, fernando2018towards, fernando2019iterative}, learning based controllers~\citep{yang2021neural, dadhich2020adaptation, azulay2021wheel, halbach2019neural, backman2021continuous}, and controllers based on heuristics~\citep{tampier2021autonomous, cardenas2023autonomous}. Many of these approaches require an accurate model of the machine and/or a model of the material being loaded, making them susceptible to modeling errors. Moreover, approaches based on data recorded by expert operators have the drawback of being limited in performance by the skill level of said operators~\citep{dadhich2016key}.

Reinforcement Learning (RL) has recently been used to address the problem of autonomous loading using front-loading earth-moving machines~\citep{dadhich2020adaptation, azulay2021wheel, backman2021continuous, shen2024generalized}. Since RL leverages interactions between an agent and its environment to derive a policy~\citep{sutton2018reinforcement}, controllers trained using RL do not require expert demonstrations. Although under certain restrictions RL-based controllers may be trained directly in the real world (e.g., as in~\cite{dadhich2020adaptation}), for earth-moving tasks, training the policies in simulation (and deploying them in the real world afterwards) avoids performing exploratory, potentially hazardous maneuvers with real industrial machinery. For these tasks however, due to the absence of a reliable tool-soil interaction model, reinforcement learning cannot guarantee an optimal performance or a successful sim2real transfer~\citep{dadhich2016key}. Previous works have addressed RL-based loading using wheel loaders~\citep{dadhich2020adaptation, azulay2021wheel} or using simulated LHDs~\citep{backman2021continuous}, but not using LHDs operating in the real world.

In this work, we propose a methodology to synthesize a deep RL-based controller to automate the loading task performed by LHDs in draw points used in block/panel caving and sublevel stoping mines. The controller is trained in a simulated environment using the Deep Deterministic Policy Gradient (DDPG) algorithm, a model-free, actor-critic algorithm proposed by \cite{lillicrap2015continuous}. To simulate a 3D muck pile, we use the soil model proposed by~\cite{luengo1998modeling}, with the addition of voxel modeling. Moreover, the controller is trained to fulfill several operational requirements (as described in \cite{dadhich2016key} and \cite{tampier2021autonomous}), for instance, minimizing wheel skidding and the time it takes to perform the loading task, and filling the bucket with a certain amount of material.

In order to validate the proposed approach, and, in particular, to test whether the  developed simulation allows addressing the sim2real transfer problem, the learned controller is deployed in a real-world scenario where a scaled LHD has to load material from a scaled draw point, similar to those found in block caving. The performance of the controller is compared to the performance of human operators (controlling the scaled LHD via teleoperation) and to the performance of the heuristic controller proposed by~\cite{tampier2021autonomous}.

With the above, the main contributions of this paper are the following:
\begin{itemize}
    \item A methodology to synthesize a deep RL-based controller to automate the loading task performed by LHDs in draw points present in block caving and sublevel stoping mines.
    \item The development of a fast, simplified simulation based on a classical soil model and the utilization of voxels for the computation of the main forces involved in the interaction between the LHD and the muck pile, during the loading process.
    \item An empirical validation of the learned controlled in the real world through a direct sim2real transfer, where the controller is deployed in a scaled LHD to load material from a scaled muck pile, outperforming other control strategies in terms of several performance metrics relevant to the loading task.
\end{itemize}

\section{Related work}
\label{sec:related_work}

The autonomous loading problem using LHDs has been addressed in the literature by means of multiple control strategies, which, in general, can be divided into five main categories. 

The first category corresponds to bucket trajectory controllers, which attempt to control the LHD's bucket so that it follows a predefined trajectory (e.g. \cite{Filla2017TowardsFT, chen2024shovel, Chen2023MachineLS, chen2022research, meng2019bucket, cao2019skid}). The trajectories for the bucket are designed specifically for the conditions of the muck pile, and are based on the knowledge of expert operators. Deep learning strategies have been used to optimize both the trajectory for loading material~\citep{Chen2023MachineLS} and the controller's parameters~\citep{chen2024shovel}. For these controllers, it is difficult to ensure a precise tracking of the bucket trajectory during the loading execution, especially in cases such as loading from muck piles with fragmented rocks~\citep{dadhich2016key}. 

The second category corresponds to force and admittance controllers, which evaluate the resistance exerted by the muck pile and output commands for the machine (e.g. \cite{marshall2008toward, dobson2017admittance, fernando2018towards, fernando2019iterative, Aoshima2021SimulationBasedOO}). In~\cite{marshall2008toward} it is proposed that an admittance controller would overcome the shortcomings of pure trajectory control and an example is given on how to design an admittance controller using data extracted from loading attempts performed by expert operators. The controller proposed by~\cite{fernando2019iterative} is an example of an admittance controller augmented with iterative learning to update  parameters to account for different types of material. 

The third category corresponds to heuristic controllers based on knowledge from expert LHD operators (e.g.~\cite{tampier2021autonomous, cardenas2023autonomous}). These controllers check the state of the machine and, according to said state, execute a predefined action.

The fourth category corresponds to controllers that are learned given a data set of expert demonstrations. These controllers are often parameterized by artificial neural networks, and trained using behavioral cloning (e.g.~\cite{halbach2019neural, yang2020learning, yang2021neural}). Since these controllers rely on the quality of the data set they are trained on, their performance is bound by the performance of the expert generating the demonstrations (which may use a sub optimal strategy to perform the loading task).

The fifth category corresponds to controllers trained using Reinforcement Learning (RL) or deep RL for the autonomous loading task performed by front-loading earth-moving machines. Using RL allows synthesizing controllers without requiring prior expert demonstrations, but also to further improve controllers that have been trained to imitate expert behaviors. In the literature, \cite{backman2021continuous} has addressed autonomous loading for LHDs using deep RL, however, the obtained policies were only tested in a simulated environment.
 
In \cite{dadhich2020adaptation}, an autonomous loading controller for a wheel-loader, trained using expert demonstrations, is adapted to a different pile environment using he Deterministic Policy Gradient (DPG) algorithm~\citep{silver2014deterministic}. The controller shows improvements in the amount of material loaded by applying reinforcement learning using few real-world trials. In \cite{shen2024generalized}, an autonomous loading controller (also for a wheel loader) is trained in simulation by combining behavioral cloning and RL, and compared to a pure RL-based controller trained under the same conditions. This work shows that the controller that leverages expert demonstrations achieves a better performance than the controller that is trained using pure RL.

While the controllers described above have been successful in performing the autonomous loading task, they rely on expert demonstrations (whether to initialize the controller or to side-step the exploration problem that would be encountered by learning from scratch), and are designed for non-mining applications using wheel loaders. 

In \cite{azulay2021wheel}, an autonomous loading controller for a real custom-made wheel loader is  obtained using pure RL (using the Deep Deterministic Policy Gradient algorithm), conducting the training process entirely in simulation, where the material to be loaded is represented as a set of particles. In \cite{backman2021continuous}, on the other hand, two RL agents are trained using the Soft Actor Critic (SAC) algorithm~\citep{haarnoja2018soft} to perform autonomous loading using LHDs. Once per load, one agent decides the best location in the pile from which to remove material, and the second agent performs the loading maneuver at the selected location. The position selection agent is added in this work so as to avoid cases where the loading agent is unable to perform the maneuver.

In this work we propose a deep RL-based controller for loading piles of blasted rocks using LHDs. The proposed method does not use any expert demonstration, and while it is trained entirely in simulation, is able to perform adequately in the real world without any major fine-tuning. As in \cite{egli2022soil} and \cite{egli2024reinforcement}, where controllers for autonomous excavation are obtained using deep RL (also conducting the training process in simulation), we leverage the Fundamental Equation of Earth-Moving so as to achieve a low computational cost (compared to accurately simulating the machine-material interactions).

\section{Proposed approach}
\label{sec:proposed_approach}

\subsection{Problem formulation}

In this work, we address the loading stage of the \textit{V-cycle}~\citep{filla2014study}, also called the excavation stage. This stage starts when the LHD hits the muck pile at maximum speed and ends when the bucket is full and the machine can pull back from the pile~\citep{tampier2021autonomous}. To address this task using RL, the interaction between the LHD (the agent) and the muck pile is modeled as a Partially Observable Markov Decision Process (POMDP), which is defined by a set of states $\mathcal{S}$, a set of actions $\mathcal{A}$, a reward function $\mathcal{R}(s,a)$, a transition function $\mathcal{T}$, a set of observations $\Omega$, an observation function $\mathcal{O}$, and a discount factor $\gamma \in [0,1)$. At each discrete time step $t$, the agent, being in a state $s_t\in\mathcal{S}$, selects and executes an action $a_t\in \mathcal{A}$ according to its policy $\pi(a_t|o_t)$, where $o_t\in \Omega$. As a consequence, the agent receives a scalar reward $r_t$ and transitions to a new state $s_{t+1}$. Under this framework, the goal is to learn a policy (that is, a controller), such that the agent maximizes the expected discounted sum of rewards that it gets by interacting with the environment. The above can be stated as the maximization of~\eqref{eq:rl_target}.
\begin{equation} 
\label{eq:rl_target}
J_{\text{RL}}(\pi)=\mathbb{E}_{a_t\sim\pi(a_t|o_t)}\left[\sum^{T}_{t=1}\gamma^{t-1}r_t\right]
\end{equation}

The interaction between the LHD machine and its environment is simulated to generate the experiences required for learning a policy. To generate said experiences, the LHD attempts to perform loading maneuvers, which end whenever the agent reaches a terminal state (thus, the time step $T$ is finite in the definition of  $J_{\text{RL}}(\pi)$). The observations and actions are constructed considering the sensors and constraints of an LHD machine, while the reward function is designed so as to guide the learning process, encouraging desirable behaviors and the fulfillment of certain operational constraints. 

In what follows, the simulation of the agent-environment interactions, and the formulation of the loading task performed by the LHD machine as a POMDP, are described in detail.

\subsection{Simulation environment}
\label{sec:sim_env}

The agent-environment interaction is simulated in Gazebo~\citep{koenig2004design}. A custom plugin was developed to calculate the total force exerted by the muck pile on the bucket of the LHD, and to apply said force in the simulation. This force is calculated using the extended Fundamental Earth-moving Equation (FEE), proposed by~\cite{luengo1998modeling}. Using an analytical equation to calculate the force exerted on the LHD's bucket (compared to actually simulating granular material and its interaction with the machine), lowers the computational cost of the simulation, which speeds up the policy's training process. 
\begin{figure}
    \centering
    \includegraphics[width=0.9\linewidth]{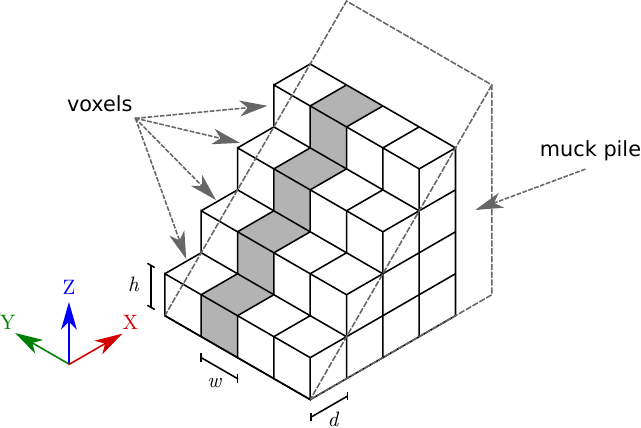}
    \caption{Diagram of the muck pile division into voxels. Each column of voxels (one of them marked in gray) has a width $w$.}
    \label{fig:muck_pile_3d}
\end{figure}

It is worth noting that the FEE allows the computation of the force exerted on the bucket in two dimensions. To apply the FEE in three dimensions, the muck pile is divided into voxels. To voxelize the muck pile, its width, height and depth are divided into $n_y$, $n_z$, and $n_x$ slices in the $y$, $z$, and $x$-axis, respectively. The resulting voxels have width $w$, height $h$, and depth $d$ (see Fig.~\ref{fig:muck_pile_3d}). Each group of vertically aligned voxels (along the $\text{XZ}$ plane) will be referred to as a ``columns'' from now on. For each column, the corresponding force exerted by the muck pile on the bucket during the loading process is calculated using the FEE. The total force on the bucket is computed as the sum of forces considering all the columns.

\begin{figure}
    \centering
    \includegraphics[width=0.9\linewidth]{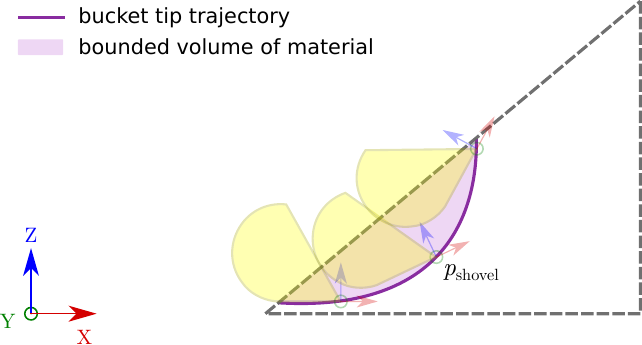}
    \caption{Illustration to represent the side view of the muck pile, and the ideal volume of material that would be loaded if the bucket tip would follow a certain trajectory.}
    \label{fig:loaded_material_diagram}
\end{figure}

Dividing the muck pile into voxels allows adding variability to the simulation of material, as each voxel can be configured with a different set of soil parameters. For instance, a voxel with a high density may simulate the presence of a large-sized rock in a given column, as it would increase the force exerted by the pile on the LHD's bucket (computed using the FEE). 

Fig.~\ref{fig:loaded_material_diagram} shows an illustration of an arbitrary trajectory that the bucket tip may follow when attempting to load from a muck pile. We determined by trial and error that the full amount of material bounded by a given trajectory cannot be considered all as loaded because the simulation used does not consider material flow. The amount of loaded material is therefore computed as a fraction of the ideal volume bounded by the bucket tip trajectory, and said fraction is set to 2/3.

\subsection{The loading task as a POMDP}

\subsubsection{Episodic settings}
\label{subsubsec:episodic_settings}

The loading task, as modeled in this work, is episodic. A given episode starts after the LHD's bucket collides with the muck pile, and ends either as a successful or unsuccessful loading attempt, depending on various conditions. 

For each episode, a new muck pile is generated with a different set of soil parameters, a fixed height and width, $h_\text{pile}$ and $w_\text{pile}$, and a different slope $\alpha \in [\alpha_\text{min}, \alpha_{\text{max}}]$ (which makes the pile's depth vary). The machine is set in an ``attack pose'' and placed at a certain distance $d_\text{attack}$ from the the muck pile's origin, so that it can reach a linear ``attack speed'' within a predetermined range. The attack pose is a configuration for the LHD arm in which the bucket is at ground level and its tip's frame $x$-axis is also parallel to the ground (see for example the leftmost bucket depiction in Fig.~\ref{fig:loaded_material_diagram}). The attack speed is the speed (in the $x$-axis) at which the machine hits the muck pile, and is different for each episode. An episode begins when the tip of the bucket, $p_\text{shovel}$ (see Fig.~\ref{fig:loaded_material_diagram}) is a distance $d_\text{init}$ inside the muck pile.

The initial settings of an episode take into consideration that the system may be integrated as a module in a larger automation stack. In particular, it is assumed that when the loading task starts, the LHD is aligned to the muck pile from which material is to be loaded. This assumption can be easily fulfilled given the availability of an autonomous navigation system and a system for muck pile detection, such as those presented in~\cite{cardenas2023autonomous}.

\begin{figure}
    \centering
    \includegraphics[width=0.9\linewidth]{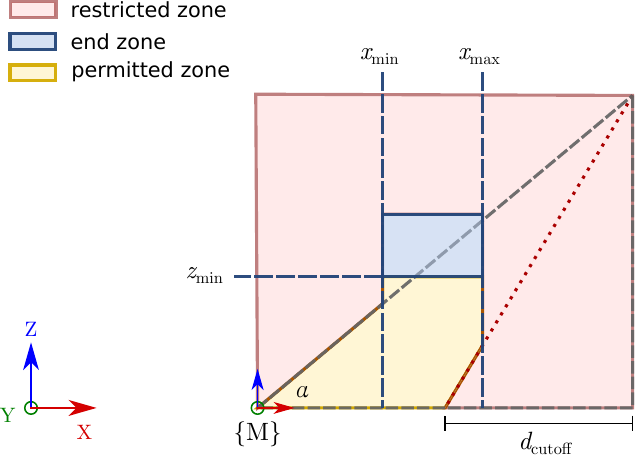}
    \caption{Division of the muck pile profile into zones in the XZ plane. The blue zone corresponds to the ``\textit{end zone}'', the yellow zone to the ``permitted zone'', and the red zone to the ``restricted zone''. All these zones are referenced to the $\{\text{M}\}$ frame.}
    \label{fig:muck_pile_zones}
\end{figure}

Determining whether a loading attempt is successful or not is more complex, and takes into account expert knowledge. To do the aforementioned, the extraction point is divided into three zones in the XZ plane: (i) an \textit{end zone}, (ii) a permitted zone, and (iii) a restricted zone. Fig.~\ref{fig:muck_pile_zones} shows a diagram of the side view of a extraction point, where the dotted lines represent the muck pile profile (with an associated frame $\{\text{M}\}$), and each listed zone is colored differently.

These zones are utilized to check where the LHD's bucket is during the execution of the loading task. To do so, the position $p^t_{\text{shovel}}=(x^t_{\text{shovel}}, y^t_\text{shovel}, z^t_\text{shovel})$ is always tracked in the XZ plane. If $p^t_{\text{shovel}}$ reaches the \textit{end zone} (i.e. $(x_\text{min}<x^t_\text{shovel}<x_\text{max}) \land (z^t_\text{shovel}>z_\text{min})$), then the episode is terminated and considered as successful. For the \textit{end zone}, $z_\text{min}$ is fixed and set according to the LHD's arm kinematics as the height at which the bucket tip should be above after a loading trial ends. There is no explicit $z_\text{max}$ coordinate for the \textit{end zone} as said coordinate is irrelevant to assess if a given loading trial has been successful.\footnote{Since an episode starts with the LHD burying the bucket into the muck pile from a starting ``attack pose'', this implies that the bucket is at ground level at the beginning of an episode. On the other hand, an episode ends if the bucket leaves the permitted zone or reaches the end zone. Given the geometry of the muck pile and the position of the end zone on it (see Fig.~\ref{fig:muck_pile_zones}), these conditions imply that there is no need to define an upper height limit for the end zone.} The parameters $x_\text{min}$ and $x_\text{max}$, on the other hand, are set according to the mass of all the material available until their respective coordinates, that is, until $x_\text{min}$ there is a mass of material $m_\text{min}$, and until $x_\text{max}$, a mass of material $m_\text{max}$. Since $x_\text{min}$ and $x_\text{max}$ depend on the available material in the muck pile, they vary as the slope $\alpha$ varies. 

Removing the bucket from the permitted zone into the restricted zone signifies that the episode ends unsuccessfully. Note that since the bucket should not reach depths greater than a certain maximum distance, a $d_\text{cutoff}$ distance, measured from the end of the muck pile, is utilized to geometrically constrain the permitted zone (as illustrated in Fig.~\ref{fig:muck_pile_zones}).

Finally, episodes are also terminated after a timeout (i.e., after a fixed amount of time steps pass), although this condition is rather artificial and does not imply the agent reaches a terminal state.

\subsubsection{Observations and actions}
\label{subsubsec:obs_and_acts}

\begin{figure}
    \centering
    \includegraphics[width=0.9\linewidth]{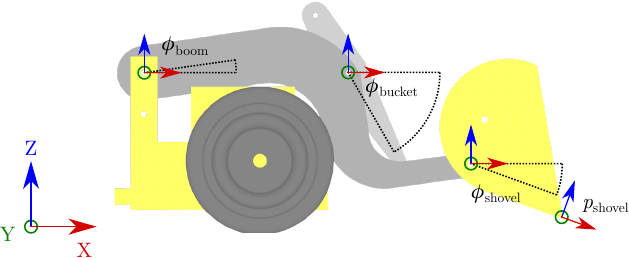}
    \caption{Diagram of the LHD's arm joint angles and their references, $\phi_{\text{boom}}$ and $\phi_{\text{bucket}}$, and the bucket tip (shovel) position $p_{\text{shovel}} = (x_{\text{shovel}}, y_{\text{shovel}}, z_{\text{shovel}})$ and pitch angle $\phi_{\text{shovel}}$.}
    \label{fig:lhd_angles_diagram}
\end{figure}

\begin{figure}
    \centering
    \includegraphics[width=0.9\linewidth]{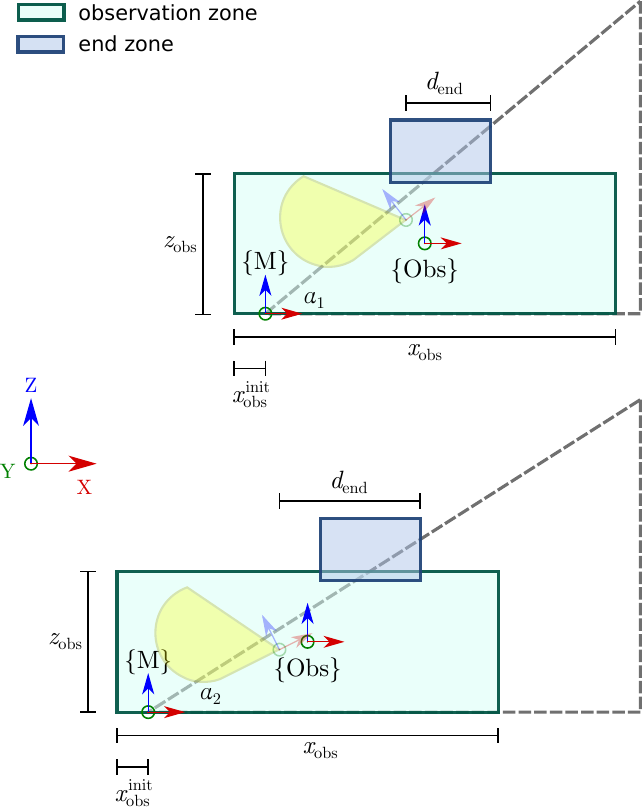}
    \caption{Illustration for the \textit{observation zone} and \textit{end zone} for two muck piles with different slopes ($\alpha_1 \neq \alpha_2$). The frames $\{\text{M}\}$ and $\{\text{Obs}\}$ correspond to the muck pile and the \textit{observation zone} frames, respectively.}
    \label{fig:obs_zone}
\end{figure}

The agent's observations, $o_t$, are defined as 12-dimensional vectors constructed using the position of the bucket tip, $p_{\text{shovel}} = (x_{\text{shovel}}, y_{\text{shovel}}, z_{\text{shovel}})$, its pitch angle value, $\phi_{\text{shovel}}$, the angular position and velocity of the LHD's arm joints, $(\phi_{\text{boom}}, \Dot{\phi}_{\text{boom}}, \phi_{\text{bucket}}, \Dot{\phi}_{\text{bucket}})$, the linear speed of the LHD in the $x$-axis, $v_x$, a drift wheel indicator, $\mathbb{I}_{\text{drift}} \in \{0, 1 \}$, the slope of the muck pile, $\alpha$, and the distance $d_\text{end}$, computed between the tip of the bucket and the farthest side of the ``\textit{end zone}'' (as illustrated by Fig.~\ref{fig:obs_zone}, $d_\text{end}$ is computed as the distance between $x_\text{shovel}$ and $x_\text{max}$). Both the LHD's bucket tip position and arm angles are illustrated in Fig.~\ref{fig:lhd_angles_diagram}. Note that the subscript ``shovel'' refers to the bucket tip and was chosen to differentiate it from the subscript ``bucket'', which refers to the bucket joint. 

To construct the observations, we also define an ``\textit{observation zone}'', which is used to map the values of the the bucket tip position, $(x_{\text{shovel}}, y_{\text{shovel}}, z_{\text{shovel}})$ to the $[-1, 1]$ range. Since different muck piles have different slopes, their starting position (relative to their maximum depth) changes; the starting position of a given muck pile determines the position of the \textit{observation zone} in the $x$-axis (given by an offset $x^\text{init}_\text{obs}$), however, its dimensions, $x_\text{obs}$, $y_\text{obs}$ and $z_\text{obs}$ remain constant. Moreover, both the position and size of the end zone are dynamic and change once at the start of every episode depending on the slope of the muck pile. The above is illustrated in Fig.~\ref{fig:obs_zone} (note that the $y$-axis dimension is not explicit in the illustration). 

The linear speed of the LHD in the $x$-axis, $v_x$, is mapped to the $[-1, 1]$ range considering the maximum speed that the machine can attain. The slope of the muck pile, $\alpha$, is mapped to $[0, 1]$ considering a minimum slope of $\alpha_\text{min}$ and a maximum slope of $\alpha_\text{max}$. The distance between the tip of the bucket and the farthest side of the end zone, $d_\text{end}$, is mapped to the $[0, 1]$ range by dividing it by the maximum possible value it may take, $d_\text{end}^{\text{max}}$, given all possible pile configurations. The drift wheel indicator, $\mathbb{I}_{\text{drift}} \in \{0, 1 \}$, takes a value of 1 when any wheel of the LHD is drifting, and is 0 otherwise. Wheel drift is detected in simulation by comparing the wheel speeds and the total movement of the machine, whilst in the real world, besides the above, it is detected by also analyzing the power usage of the wheel motors.

The angular position and velocity of the LHD's arm joints $(\phi_{\text{boom}}, \Dot{\phi}_{\text{boom}}, \phi_{\text{bucket}}, \Dot{\phi}_{\text{bucket}})$ along with the the pitch angle value $\phi_{\text{shovel}}$ are obtained from different frames in the LHD's arm, which are shown in Fig.~\ref{fig:lhd_angles_diagram}. Each of these observations is mapped to $[-1, 1]$ considering the full range of motion and the speed limits of the real machine.

For the agent's actions, $a_t$, we consider that during the loading stage the machine can be controlled by three different commands: $u_\text{wheels}$, a speed command for the wheels, $u_\text{boom}$, a speed command for the boom joint, and $u_\text{bucket}$, a speed command for the bucket joint. Because the LHD has four-wheel drive, all the wheels perform the same action given a $u_\text{wheels}$ command. Note that the LHD's central articulation joint is not used during the loading process because mine operators' experience indicate that loading with a non-straight central articulation may cause damage to its actuator. 

With the above, the agent's actions are defined as 3-dimensional vectors $a_t=(a^t_\text{wheels}, a^t_\text{boom}, a^t_\text{bucket})$, where each action's component is in the $[-1, 1]$ range and then mapped to the speed range of its respective actuator. The actuation of the LHD given the commands resulting from the denormalization of $a_t$ is constrained to a desired range of motion that avoids behaviors that do not contribute to a successful loading. In the case of the wheels, $u_\text{wheels}\in [v_\text{wheels}^\text{min}, v_\text{wheels}^\text{max}]$, and $v_\text{wheels}^\text{min}=0$, so that the machine cannot go backwards, because the loading is intended to be a single (forward) movement, and going backwards would end the maneuver promptly. In the case of the boom and bucket joints, $u_\text{boom}\in [\Dot{\phi}_\text{boom}^\text{min},\Dot{\phi}_\text{boom}^\text{max}]$, and $u_\text{bucket}\in [\Dot{\phi}_\text{bucket}^\text{min},\Dot{\phi}_\text{bucket}^\text{max}]$, where  $\Dot{\phi}_\text{boom}^\text{min}<0$, $\Dot{\phi}_\text{boom}^\text{max} =0$, $\Dot{\phi}_\text{bucket}^\text{min}=0$, and $\Dot{\phi}_\text{bucket}^\text{max}>0$, that is, the boom cannot be lowered and the bucket cannot be tilted forward to avoid dangerous situations (such as lifting the front wheel axis), which may damage the machine. 

All the observations and actions that have been described in this section are summarized in Table~\ref{tab:obs_and_acts}.

\begin{table}
\centering
\caption{Description of the components of the agent's observations and actions.}
\begin{tabular*}{\linewidth}{cl@{\extracolsep{\fill}}l}
\toprule
& \textbf{Component} & \textbf{Description} \\ 
\midrule
\multirow{8}{*}{\rotatebox{90}{Observations}}& $(x_{\text{shovel}}, y_{\text{shovel}}, z_{\text{shovel}})$                            & Position of the bucket tip \\
& $\phi_{\text{shovel}}$        & Pitch angle of the bucket tip  \\ 
& $v_x$                                     & LHD linear speed  \\
& $\mathbb{I}_{\text{drift}}$                 & Drift indicator  \\
& $(\phi_{\text{boom}}, \Dot{\phi}_{\text{boom}})$ & \textit{boom} angular position and velocity \\
& $(\phi_{\text{bucket}}, \Dot{\phi}_{\text{bucket}})$  & \textit{bucket} angular position and velocity \\
& $d_\text{end}$                              & $|x_\text{shovel} - x_\text{max}|$ \\ 
& $\alpha$                                    & Slope of the muck pile\\ 
\midrule
\multirow{3}{*}{\rotatebox{90}{Actions}}& $u_\text{wheels}$                           & Wheels speed command          \\
& $u_\text{boom}$                             & Boom joint speed command      \\
& $u_\text{bucket}$                           & Bucket joint speed command    \\ 
\bottomrule
\end{tabular*}
\label{tab:obs_and_acts}
\end{table}

\subsubsection{Reward function}

The reward function used in this work is defined by Eq.~\eqref{eq:r_total}. This function is designed to guide the agent towards learning a policy that allows loading as much material as possible while minimizing the execution time of the task and the risk of damaging the LHD machine. 
\begin{multline} 
\label{eq:r_total}
    r_\text{t} = r_\text{traj}^t + r_\text{midgoal}^t + r_\text{inact}^t  + r_\text{drift}^t + r_\text{stuck}^t + r_\text{dump}^t  \\ + r_\text{bottom}^t 
 + r_\text{weight}^t + r_\text{success}^T + r_\text{zone}^T
\end{multline}

Note that this function is defined as the sum of multiple components, which have a superscript $t$ if they are computed at each time step, or a superscript $T$ if they are computed only when the agent reaches a terminal state. In what follows, each of these components is described.

\paragraph{Shovel trajectory reward $(r_\text{\normalfont traj}^t)$}

This reward component is designed to encourage, but not to specify, desirable trajectories for the bucket tip while it is inside the muck pile. To compute $r_\text{traj}^t$, an estimation of the instantaneous bucket tip trajectory in the XZ plane is bounded so as to dynamically restrict the actions that the agent may take.
 
To represent the instantaneous trajectory of the bucket tip, the 2D vector defined by the displacement of the bucket tip between $t-1$ and $t$ in the XZ plane is computed. Thus, if the vector defined by $(x^{t-1}_{\text{shovel}}- x_{\text{shovel}}^{t}, z^{t-1}_{\text{shovel}} -  z_{\text{shovel}}^{t})$ and referenced in the bucket tip frame is out of bounds, then the agent is penalized.  

To upper bound the vector in the $z$-axis, we consider an auxiliary point, $(x_{\text{min}}, z_{\text{upper}})$, where $z_{\text{upper}}$ is half the height of the muck pile when $x^{t}_\text{shovel}<x_{\text{min}}$ and is equal to $z_{\text{min}}$ when $x^{t}_\text{shovel}\geq x_{\text{min}}$. With the above, the upper bound is defined as the vector between the current shovel tip position in the XZ plane and the aforementioned auxiliary point, i.e., as $(x_{\text{min}}- x^{t}_\text{shovel}, z_{\text{upper}}- z^t_{\text{shovel}})$.

For the lower bound, we define an auxiliary curve in the XZ plane which will be referred to as ``rail''. This rail has an exponential shape, and is defined by Eq.~\eqref{eq:rail}, where $x_\text{offset}$, $x_{\text{max}}$ and $z_\text{offset}$ determine the position of the rail in the XZ plane, and $x_\Delta$ defines how far ahead from the bucket tip position (in the $x$-axis) a point on the rail curve is going to be computed. Thus, the lower bound simply corresponds to the vector between the tip of the bucket in the XZ plane, and $(x^{t}_\text{shovel}+x_\Delta, z^{t}_\text{rail})$, where $z^t_\text{rail}$ is computed using \eqref{eq:rail}.\footnote{The exponent in the definition for $z_\text{rail}$ is arbitrary, in the sense that exponents around 17 would produce curves that are also usable to compute the lower bound vector.} 
\begin{equation}
    z^{t}_\text{rail}(x^{t}_{\text{shovel}}) = (x^{t}_\text{shovel} + x_\Delta + (x_\text{offset} - x_\text{max}))^{17} - z_\text{offset}
    \label{eq:rail}
\end{equation}

\begin{figure}
    \centering
    \includegraphics[width=0.9\linewidth]{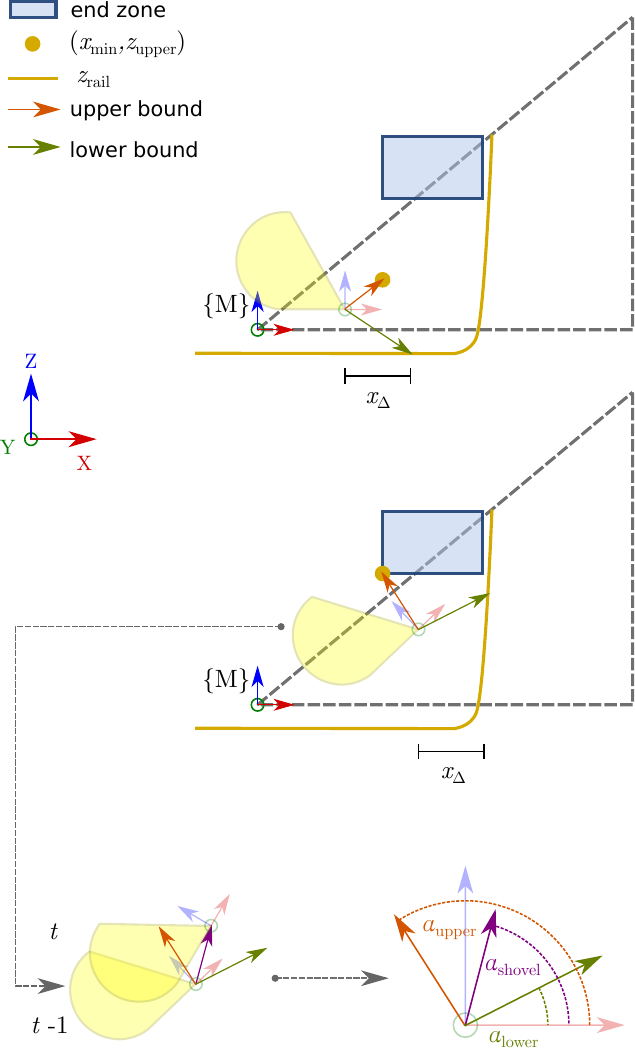}
    \caption{The muck pile at the top shows how the vectors associated with the upper and lower bounds for the bucket tip trajectory are computed when $x^t_\text{shovel}<x_\text{min}$, whereas the muck pile below shows the same for $x^t_\text{shovel}\geq x_\text{min}$. At the bottom, it is shown the way in which the vector associated with $\alpha_\text{shovel}$ is computed, and how $\alpha_\text{upper}$, $\alpha_\text{lower}$, and $\alpha_\text{shovel}$ are compared.}
    \label{fig:rew_traj}
\end{figure}

With the above definitions, the shovel trajectory reward, $r^{t}_{\text{traj}}$, is given by Eq.~\eqref{eq:r_splines}, where $R_\text{traj} >0$, $\alpha_\text{shovel}$ is the angle associated with the bucket tip trajectory estimate, $\alpha_\text{lower}$ is the angle associated to the lower bound vector, and $\alpha_\text{upper}$ is the angle associated to the upper bound vector. Fig.~\ref{fig:rew_traj} illustrates the variables utilized to compute this reward component.
\begin{equation}
       r_{\text{traj}}^t = -R_{\text{traj}} \cdot \mathds{1}_{\{(\alpha_{\text{shovel}}<\alpha_{\text{lower}}) \lor (\alpha_{\text{shovel}}>\alpha_{\text{upper}})\} }
    \label{eq:r_splines}
\end{equation}

Note that the upper bound guides the agent to first move forward without raising the bucket tip, and after passing $x_{\text{min}}$ in the $x$-axis, to raise it so as to reach the \textit{end zone}. The lower bound, on the other hand, guides the agent to raise the bucket when approaching $x_{\text{max}}$. Both of these bounds aim at allowing the bucket tip position to reach the \textit{end zone}, which is crucial to end an episode successfully.

\paragraph{Mid goal reward $(r_\text{\normalfont midgoal}^t)$}
\label{sec:midgoal}

Near the starting point of the muck pile, the agent may prematurely end the loading task by lifting the bucket prior to loading a sufficient amount of material. The $r_\text{midgoal}^t$ component is designed to prevent the above from occurring, and is defined by Eq.~\eqref{eq:r_carrot}, where $x_\text{midgoal}$ is an arbitrary constant depth within the extraction point along the $x$-axis, $d^{t}_\text{midgoal} = |x^t_{\text{shovel}} - x_\text{midgoal}|$, $d_\text{midgoal}^{\text{max}}$ is the maximum value for $d^{t}_\text{midgoal}$, and $R_\text{midgoal} > 0$ is a constant value.
\begin{equation}
       r_{\text{midgoal}}^t = -R_{\text{midgoal}} \cdot \frac{d^{t}_\text{midgoal}}{d^{\text{max}}_\text{midgoal}}\cdot \mathds{1}_{\{ x^{t}_\text{shovel} < x_{\text{midgoal}}\} }
    \label{eq:r_carrot}
\end{equation}

\begin{figure}
    \centering
    \includegraphics[width=0.9\linewidth]{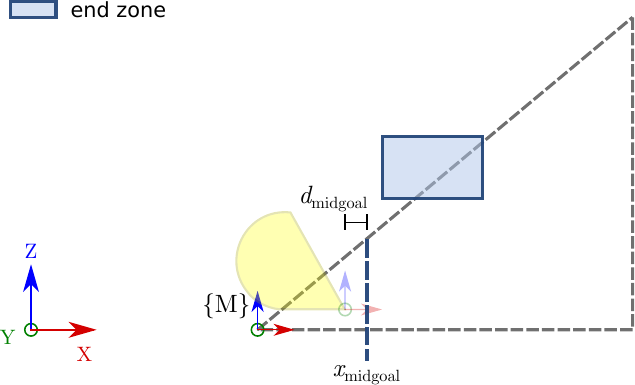}
    \caption{The mid goal reward diagram incentivizes the agent to bury the bucket inside the muck pile. This reward component takes the value of zero once the tip of the bucket exceeds the value of $x_{\text{midgoal}}$.}
    \label{fig:rew_midgoal}
\end{figure}

Note that $r_\text{midgoal}^t$ constantly penalizes the agent until $x_\text{shovel} \geq x_{\text{mid\_goal}}$, which guides it towards burying the bucket into the pile a certain distance in the $x$-axis before starting to lift it. Fig.~\ref{fig:rew_midgoal} shows a diagram of the parameters for this reward component.

\paragraph{Inactivity reward $(r_\text{\normalfont inact}^t)$}

This reward component is designed to penalize the agent if it does not execute actions to lift the bucket tip with a magnitude greater than a given threshold. This component can only be non-zero after the bucket tip has exceeded the target depth $x_\text{midgoal}$, since forcing any boom or bucket action at the start of the excavation may end the loading maneuver prematurely. Given the above, $r^t_\text{inact}$ is given by Eq.~\eqref{eq:r_inact}, where  $u^{\text{thresh}}_{\text{boom}}$ and $u^{\text{thresh}}_{\text{bucket}}$ are threshold values for the boom and bucket speed commands, respectively, and $R_\text{inact}>0$ is a constant value.
\begin{equation}
       r_{\text{inact}}^t = -R_{\text{inact}} \cdot \mathds{1}_{\left\{\left(\left|\frac{u^{t}_{\text{boom}}}{\Dot{\phi}^\text{min}_\text{boom}}\right|< u^{\text{thresh}}_{\text{boom}} \lor \left|\frac{u^{t}_{\text{bucket}}}{\Dot{\phi}_\text{bucket}^{\text{max}}}\right|< u^{\text{thresh}}_{\text{bucket}}\right) \land (x^t_{\text{shovel}}\geq x_{\text{midgoal}})\right\} }
    \label{eq:r_inact}
\end{equation}

\paragraph{Drift reward $(r_\text{\normalfont drift}^t)$}
This reward component encourages actions that mitigate wheel drift when said drift happens. If there is wheel drift ($\mathbb{I}_{\text{drift}}$ equals one), the agent is penalized if it does not perform any corrective action against it. These corrective actions correspond to lifting the boom or the bucket to gain traction. Given the above, $r_\text{drift}^t$ is defined by~\eqref{eq:r_drift}, where $R_\text{drift} > 0$ is a constant value.
\begin{equation}
       r_{\text{drift}}^t = -R_{\text{drift}} \cdot \mathbb{I}_{\text{drift}} \cdot \mathds{1}_{\{(u^{t}_{\text{boom}}=0) \land (u^{t}_{\text{bucket}}=0)\} }
    \label{eq:r_drift}
\end{equation}

\paragraph{Stuck reward $(r_\text{\normalfont stuck}^t)$}

This reward component is designed to penalize the agent when it gets stuck. The LHD machine is considered to be stuck when the position of the bucket tip has not moved a certain distance after a number of time steps have passed. The term $r_\text{stuck}^t$ is defined by Eq.~\eqref{eq:r_stuck}, where $\Delta^t_\text{shovel}$ is the variation of the bucket tip position in the $x$-axis between time steps $t$ and $t-1$, $\Delta_\text{thresh}$ is a fixed threshold distance, and both $R_\text{stuck}>0$ and $T_\text{stuck}>0$ are constant values.
\begin{equation}
     r^{t}_{\text{stuck}} = -R_\text{stuck} \cdot \mathds{1}_{\{(\Delta^{t}_\text{shovel}<\Delta_\text{thresh}) \land ... \land (\Delta^{t-T_\text{stuck}}_\text{shovel}<\Delta_\text{thresh}) \}}
     \label{eq:r_stuck}
\end{equation}

\paragraph{Dump reward $(r_{\text{\normalfont dump}}^t)$}
This reward is designed to penalize the agent for reaching configurations in which dumping material on the ground becomes likely. In the real world, due to the configuration of the LHD's arm, when lifting the boom joint without moving the bucket joint, the bucket begins to tilt forward, causing material to fall from it. In addition, with the bucket tilted forward, it is easier for the LHD to get stuck in the pile of material. Given the above, $r_{\text{dump}}^t$ is defined by Eq.~\eqref{eq:r_dump}, where $\phi_\text{thresh}^\text{pitch}$ is a limiting angle for $\phi_\text{shovel}^{t}$, and $R_\text{dump}>0$ is a constant value.
\begin{equation}
     r^{t}_{\text{dump}} = -R_\text{dump} \cdot \mathds{1}_{\{ \phi^{t}_\text{shovel} \geq \phi_\text{thresh}^\text{pitch}\}}
     \label{eq:r_dump}
\end{equation}

\paragraph{Bucket bottom reward $(r_\text{\normalfont bottom}^t)$}

\begin{figure}
    \centering
    \includegraphics[width=0.9\linewidth]{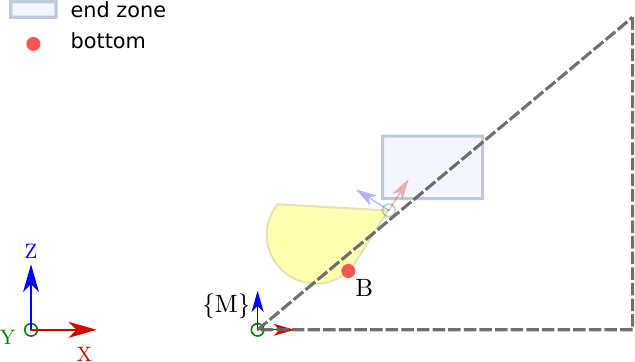}
    \caption{Example diagram for a situation where the bottom of the bucket (red dot B) is buried inside the muck pile. When the bottom of the pile is buried inside the muck pile the bucket is compressing material, which can lead to complications such as lifting the LHD's front axis.}
    \label{fig:rew_bottom}
\end{figure}

The simulation utilized in this work makes it possible for the bucket to reach poses inside the muck pile that it would not reach in the real world. This is because the simulation utilized has no contact physics, and the resistive force on the bucket is determined solely by the pose of its tip. An undesired consequence of the above, is that the bucket bottom (illustrated in Fig.~\ref{fig:rew_bottom} as the red dot labeled as ``B''), may end up buried in the muck pile, with no forces being applied on it. In the real world, burying the bottom of the bucket implies pressing the bucket against the muck pile, which may lead to lifting the front axis of the LHD, thus, damaging the machine. The $r_\text{bottom}^t$ reward component, defined in Eq.~\eqref{eq:r_bottom}, is designed to account for the simulation deficiencies by penalizing the agent if it buries the bottom of the bucket into the muck pile. In  Eq.~\eqref{eq:r_bottom}, $(x^t_\text{bottom}, y^t_\text{bottom}, z^t_\text{bottom})$ is the instantaneous position of the bottom of the bucket, $z_{\text{pile}}(x)$ is a function that returns the height of the muck pile for a given coordinate $x$ in the $x$-axis, and $R_\text{bottom}>0$ is a constant value.
\begin{equation}
     r^{t}_{\text{bottom}} = -R_\text{bottom} \cdot \mathds{1}_{\{ z^t_\text{bottom} \leq z_{\text{pile}}(x^t_\text{bottom})\}}
     \label{eq:r_bottom}
\end{equation}

\paragraph{Weight reward $(r_{\text{\normalfont weight}}^t)$}
This reward component is designed to incentivize the agent to load more material, and is defined by Eq.~\eqref{eq:r_weight}, where $R_\text{weight} > 0 $, and $W^t$ is the amount of material inside the bucket at time step $t$. Note that $R_\text{weight}$ controls the numerical range of $r_{\text{weight}}^t$ as a scaling factor.
\begin{equation}
    r_\text{weight}^{t} = R_\text{weight} \cdot W^t
    \label{eq:r_weight}
\end{equation}

\paragraph{Success reward $(r_{\text{\normalfont success}}^T)$}
\label{subsubsec:success_reward}

For a loading attempt to be considered successful, the LHD has to load as much material as possible, and to end the loading maneuver with its bucket in a pose that allows it to safely exit the muck pile, while keeping all the loaded material inside the bucket. The $r_{\text{success}}^T$ reward component is designed to incentivize the above, as it only triggers when the bucket tip position reaches the \textit{end zone} (which would allow it to get out of the  muck pile afterwards), encourages reaching this region with a certain pitch angle for the bucket tip, and rewards the agent for loading at least a certain amount of material. This component is defined by Eq.~\eqref{eq:r_bonus}, where $\Delta_\text{weight} = \Delta(W^{T}, W_\text{target}) \in [0,1]$ and $\Delta_\text{pitch} = \Delta(\phi^{T}_\text{shovel}, \phi_\text{target}) \in [0,1]$ are computed using Eqs.~\eqref{eq:eucli_log_dist} and ~\eqref{eq:eucli_single}, $W^{T}$ and $\phi^T_\text{shovel}$ correspond to the mass of the loaded material and the pitch angle of the shovel tip by the end of the excavation, $W_\text{target}$ and $\phi_\text{target}$ to their target values, $\rho(x,x_\text{target})$ to the (clipped) Euclidean distance between $x$ and $x_\text{target}$, $\rho_\text{thresh}$ to a constant threshold distance, and $R_\text{success} > 0$.
\begin{equation}
    r_\text{success}^T = R_\text{success} \cdot \frac{\Delta_\text{weight}+\Delta_\text{pitch}}{2} \cdot \max(W^T, W_\text{target}) 
    \label{eq:r_bonus}
\end{equation}
\begin{align}
    \Delta(x, x_\text{target}) &= 1 - \frac{\delta(x,x_\text{target})}{\max_x \delta(x,x_\text{target})}
    \label{eq:eucli_log_dist} \\
    \delta(x,x_\text{target}) &=
    \begin{multlined}[t]
      -(\rho(x,x_\text{target}) + \ln(\rho(x,x_\text{target}))) \\
        {} {}+(\rho_\text{thresh} + \ln(\rho_\text{thresh}))
     \end{multlined} \label{eq:eucli_single}\\
     \rho(x,x_\text{target}) &= \max\left(\rho_\text{thresh}, \|x-x_\text{target}\|_2\right)
\end{align}

Note that $\Delta(x, x_\text{target})$ is equal to zero when the distance between $x$ and $x_\text{target}$ is maximal, and takes the value of one when the distance is below a certain threshold. Given the above, $r_{\text{success}}^T$ is designed to reach its maximum value when both the final angle of the bucket and the amount of material loaded reach their target values.

\paragraph{Zone reward $(r_\text{\normalfont zone}^T)$}
The zone reward is only non zero when a terminal state indicating an unsuccessful loading attempt is reached, that is, whenever the shovel tip position enters the restricted zone (see Fig.~\ref{fig:muck_pile_zones}). This reward component is defined by \eqref{eq:r_zone}, where $\mathcal{P}_\text{restricted}$ is the set of points conforming the restricted zone, and $R_\text{zone}>0$ is a constant value.
\begin{equation}
    r_{\text{zone}}^T = -R_\text{zone} \cdot \mathds{1}_{\{p^t_\text{shovel} \in \mathcal{P}_\text{restricted}\}}
    \label{eq:r_zone}
\end{equation}

All the reward components described in this section are summarized in Table~\ref{tab:rewards}. The values of the constants used for their computation in the experiments reported in Section~\ref{sec:experimental_results} are given in Table~\ref{tab:system-params}.
\begin{table}
\centering
\caption{Reward's components summary.}
\begin{tabular*}{\linewidth}{l@{\extracolsep{\fill}}l}
\toprule
\textbf{Term} & \textbf{Definition} \\ 
\midrule

$r_{\text{traj}}^t$ & $-R_{\text{traj}} \cdot \mathds{1}_{\{(\alpha_{\text{shovel}}<\alpha_{\text{lower}}) \lor (\alpha_{\text{shovel}}>\alpha_{\text{upper}})\} }$ \\

$r_{\text{midgoal}}^t$ &  $-R_{\text{midgoal}} \cdot \frac{d^{t}_\text{midgoal}}{d^{\text{max}}_\text{midgoal}}\cdot \mathds{1}_{\{ x^{t}_\text{shovel} < x_{\text{midgoal}}\} }$ \\

$r_{\text{inact}}^t$ & $-R_{\text{inact}} \cdot \mathds{1}_{\left\{\left(\left|\frac{u^{t}_{\text{boom}}}{\Dot{\phi}^\text{min}_\text{boom}}\right|< u^{\text{thresh}}_{\text{boom}} \lor \left|\frac{u^{t}_{\text{bucket}}}{\Dot{\phi}_\text{bucket}^{\text{max}}}\right|< u^{\text{thresh}}_{\text{bucket}}\right) \land (x^t_{\text{shovel}}\geq x_{\text{midgoal}})\right\} }$ \\

$r_{\text{drift}}^t$ & $-R_{\text{drift}} \cdot \mathbb{I}_{\text{drift}} \cdot \mathds{1}_{\{(u^{t}_{\text{boom}}=0) \land (u^{t}_{\text{bucket}}=0)\} }$ \\

$r^{t}_{\text{stuck}}$ & $-R_\text{stuck} \cdot \mathds{1}_{\{(\Delta^{t}_\text{shovel}<\Delta_\text{thresh}) \land ... \land (\Delta^{t-T_\text{stuck}}_\text{shovel}<\Delta_\text{thresh}) \}}$ \\

$r^{t}_{\text{dump}}$ & $ -R_\text{dump} \cdot \mathds{1}_{\{ \phi^{t}_\text{shovel} \geq \phi_\text{thresh}^\text{pitch}\}}$ \\

$r^{t}_{\text{bottom}}$ & $-R_\text{bottom} \cdot \mathds{1}_{\{ z^t_\text{bottom} \leq z_{\text{pile}}(x^t_\text{bottom})\}}$ \\
$r_\text{weight}^{t}$ & $ \quad \!R_\text{weight} \cdot W^t$\\

$r_\text{success}^T$ & $\quad \! R_\text{success} \cdot \frac{\Delta_\text{weight}+\Delta_\text{pitch}}{2} \cdot \max(W^T, W_\text{target})$ \\

$r_{\text{zone}}^T$ & $-R_\text{zone} \cdot \mathds{1}_{\{p^t_\text{shovel} \in \mathcal{P}_\text{restricted}\}}$ \\

\bottomrule
\end{tabular*}
\label{tab:rewards}
\end{table}

\subsection{Policy parameterization and training}

\begin{figure}
    \centering
    \includegraphics[width=0.9\linewidth]{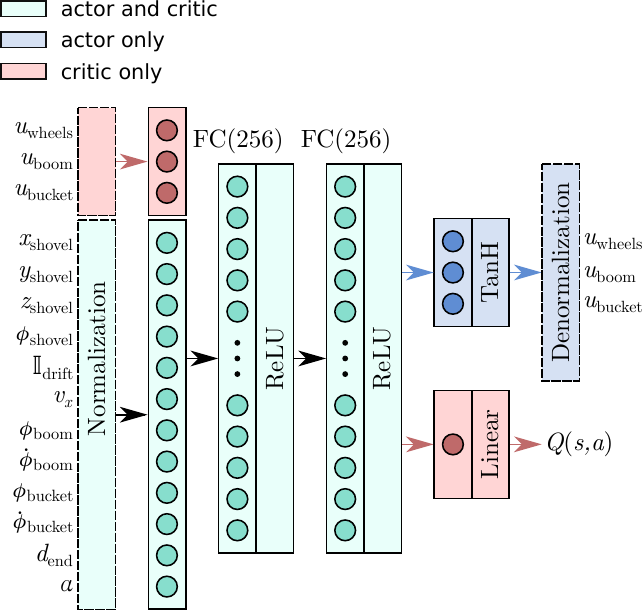}
    \caption{Actor and critic neural networks' architectures. The term ``FC($n$)'' is used to denote a \textit{fully connected} layer with $n$ units, ``ReLU'' stands for rectified linear unit, and ``TanH'' stands for hyperbolic tangent. Although the actor and critic architectures are represented in a single diagram, they do not share weights.}
    \label{fig:ddpg_parameterizations}
\end{figure}

The RL algorithm utilized in this work is the actor-critic Deep Deterministic Policy Gradient (DDPG) algorithm~\cite{lillicrap2015continuous}. Both the actor and critic are parameterized by (independent) neural networks whose architectures are shown in Fig.~\ref{fig:ddpg_parameterizations}.

All the layers of these neural networks are \textit{fully connected}. The inputs associated with observations and actions are normalized according to the procedure described in Section~\ref{subsubsec:obs_and_acts}, to construct $o_t$ and $a_t$. The actor's outputs, which are in the $[-1, 1]$ interval due to the hyperbolic tangent (TanH) used as activation function in its output layer, are denormalized before using them as commands for the LHD machine.

\section{Experimental results}
\label{sec:experimental_results}

\subsection{Experimental settings}

\begin{figure}
    \centering
    \includegraphics[width=8cm]{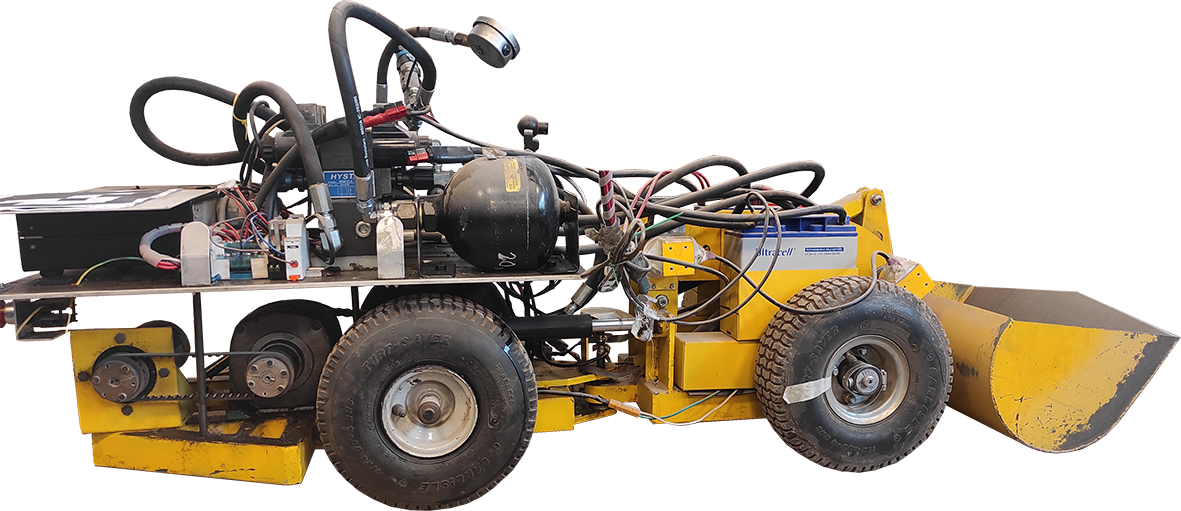}
    \caption{Side view of the scaled LHD used for real world experiments.}
    \label{fig:scaled_lhd}
\end{figure}

All the training process is conducted in simulations, using Gazebo~\citep{koenig2004design} and ROS~\citep{quigley2009ros}, however, the validation of the system is performed in the real world. All the experiments are conducted considering a custom-made, scaled LHD machine as deployment platform, which has both a simulated and a real counterpart. In both cases, a simulated and real scaled muck pile is utilized.

The real scaled-down LHD is shown in Fig.~\ref{fig:scaled_lhd}; this machine has the same degrees of freedom as a real-size LHD, it is scaled down with a ratio of 5:1, it weighs $\approx$121~kg, and its bucket is designed to hold 30~kg of material. Moreover, its articulations are actuated using an hydraulic system, and its wheels using electrical motors: the rear wheels share a single motor and are connected to a differential, whilst each front wheel has an independent motor. The machine is equipped with encoders on the arm's joints, and on all the electrical motors to obtain angular position and velocity estimations. An Arduino Due is mounted onboard of the LHD; it sends sensory information to an external computer and receives control commands for the machine. The control system for the loading task is designed to operate at a frequency of 10~Hz.

The scaled-down draw point (simulated for training, and real for evaluation) is similar to those present in block/panel caving, and also has a scale of 5:1. The material in these extraction points is subjected to large compressive forces, which in simulation are computed using the modified FEE described in Section~\ref{sec:sim_env}, and emulated in the real-world by an structure holding a column of material on top of the muck pile. The specifics of the real extraction point used in this work are described in Section~\ref{sec:real_environment_eval}.

\subsection{Training and evaluation in simulation}
\label{subsec:sim_training_and_eval}

The simulated muck pile is configured with the parameters for gravel~\citep{geo_material1,geo_material2}, and each of its voxels has an approximate size of 6~cm per side, which results in $\approx$2,100~voxels representing the whole muck pile (the parameters $n_x$, $n_y$ and $n_z$, which control the pile voxelization, are documented in Table~\ref{tab:system-params}).

During training, the conditions described in Section~\ref{subsubsec:episodic_settings} are utilized to set the initial settings of an episode: for each episode, a new muck pile is generated with different soil parameters and a different slope (varying between $\alpha_\text{min}$ and $\alpha_\text{max}$), the LHD is placed at $d_\text{attack}$ from the beginning of the pile, and a training episode begins when the tip of the bucket is a distance $d_\text{init}$ inside the muck pile. Moreover, each episode has a maximum duration of 200~time steps before a time-out occurs, and is subjected to the termination conditions that depend on the position of the shovel tip, which are also described in Section~\ref{subsubsec:episodic_settings}.

\begin{figure}
    \centering
    \includegraphics[width=\linewidth]{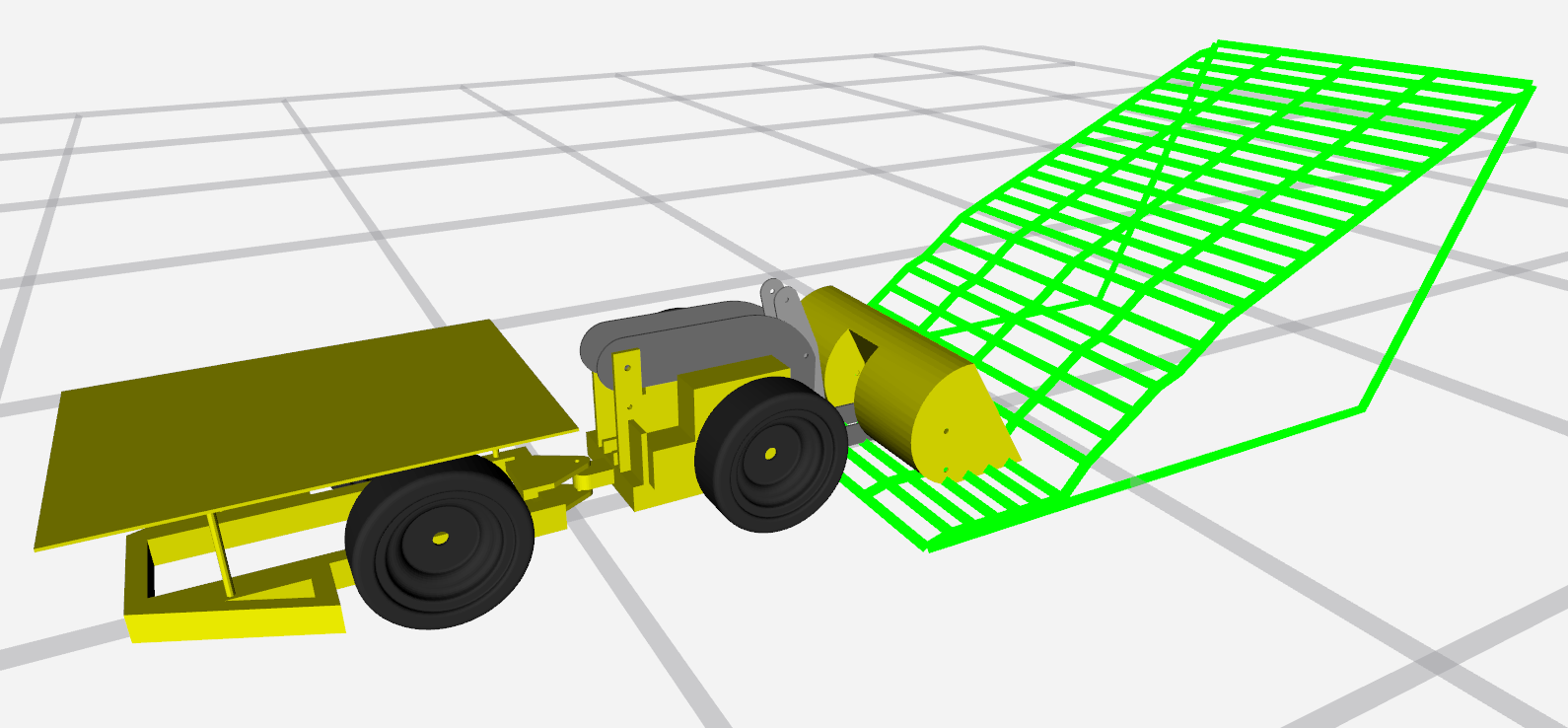}
    \caption{Simulation environment utilized to train the loading policies. The simulated muck pile is displayed using several green segments that capture its overall geometry.}
    \label{fig:scaled_lhd_sim}
\end{figure}

Fig.~\ref{fig:scaled_lhd_sim} shows the simulation environment used to train the loading policies for the scaled LHD. It is worth mentioning that the divisions in the muck pile shown in this illustration are merely utilized for visualization, and do not represent the voxels that actually divide the muck pile and that are used to compute the FEE according to what is described in Section~\ref{sec:sim_env}.

The training process is set for 300,000 time steps (a critic and policy update is performed for every simulated time step), and the exploration noise is modulated by a factor that decays linearly from 1 to 0.05 in 250,000 time steps. All the computational processing required for training is performed on a computer equipped with an AMD~Ryzen~3600 processor, and an Nvidia GeForce RTX~2060~Super graphics processing unit.

Two different types of policies are trained to solve the autonomous loading task. The first type corresponds to policies with a continuous action space, where the arm actuators and the wheels in the simulation are responsive to commands in a continuous numerical range. The second type of policies corresponds to those trained considering that the arm actuators in simulation can only execute the extreme values of the range of each action component (as in a discrete action space), however, the wheels remain responsive to continuous commands. From here on after, the first type of policies are referred to as RL-Continuous policies (RLC policies) and the second type of policies as RL-Discrete policies (RLD policies).\footnote{Note that RLD policies, despite their denomination, possess an hybrid action space (continuous and discrete).}

The LHD used for testing in the real world has ternary actuators in the arm joints, i.e., these actuators always operate at maximum speed in either direction, or do not move. The RLD policies are considered in this study to evaluate whether a policy trained with the target machine dynamics performs better than a policy trained with continuous actions. To construct the actions for the RLD policy, an action filter is utilized. Said action filter is defined by Eq.~\eqref{eq:action_filter}, where $a^\text{thresh}$ is a fixed threshold for the action component $a$. Note that this filter returns 1 if the action component surpasses $a^\text{thresh}$, and return $-$1 otherwise. After applying the filter to a given action component, the result is denormalized to get a suitable command for the LHD. The RLD policies are trained by applying this filter with $a^\text{thresh}=0$ to $a^t_\text{boom}$ and $a^t_\text{bucket}$, thus, obtaining $\hat{a}^t_\text{boom} \in \{-1, 1\}$ and $\hat{a}^t_\text{bucket} \in \{-1, 1\}$. 
\begin{equation}
    \label{eq:action_filter}
    \hat{a}(a, a^{\text{thresh}}) = \mathds{1}_{\{a > a^\text{thresh}\}} - \mathds{1}_{\{a \leq a^\text{thresh}\}}
\end{equation}

The performance evolution of the RLC and RLD policies is obtained by evaluating them in 120 loading attempts every 10,000 training steps. During this evaluation, the exploration noise is disabled, and four metrics are computed to assess the performance of the policies: \textit{Average return}, \textit{Average material loaded} and \textit{Average fill factor}. These metrics are defined as follows:

\begin{itemize}
    \item \textit{Average return} (AR): Average return (undiscounted sum of rewards) obtained by the agent during the evaluation episodes.
    \item \textit{Average material loaded} (AML): Average mass of the material loaded during the evaluation episodes.
    \item \textit{Average fill factor} (AFF): Ratio of material loaded in the bucket to the bucket capacity, averaged across evaluation episodes.\footnote{This metric is computed so as to express AML in terms of the bucket capacity.}
\end{itemize}

\begin{table}
\centering
\caption{Problem formulation parameters.}

\begin{tabular*}{\linewidth}{ll@{\extracolsep{\fill}}l}
\toprule
& \textbf{Parameter} & \textbf{Value} \\ 

\midrule
\multirow{7}{*}{DDPG}                   & Actor learning rate & 0.0001        \\
                                        & Critic learning rate & 0.001        \\
                                        & Discount rate $\gamma$ & 0.99        \\
                                        & Experience Replay Buffer size & 200,000        \\
                                        & Minibatch size & 256        \\
                                        & Smoothing factor $\tau$       & 0.001\\
                                        & OU noise ($\mu$, $\sigma$, $\theta$, $\Delta t$) & 0, 0.4, 0.15, 0.2\\

\midrule
\multirow{11}{*}{\begin{tabular}{@{}l}
     Reward  \\
     function 
\end{tabular}}                      & $R_\text{zone}$  &  2,000  \\
                                    & $R_\text{traj}$   &  60  \\
                                    & $R_\text{drift}$  &  5   \\
                                    & $R_\text{bottom}$   &  5  \\
                                    & $R_{\text{midgoal}}$, $d_\text{midgoal}^{\text{max}}$ [m]  &    25, 0.44   \\
                                    & $R_\text{stuck}$, $T_\text{stuck}$, $\Delta_\text{thresh}$  &  5, 8, 0.005  \\
                                    & $R_{\text{inact}}$, $u^\text{thresh}_\text{boom}$, $u^\text{thresh}_\text{bucket}$  &  50, 0.5, 0.5  \\
                                    & $R_\text{weight}$ & 1  \\
                                    & $R_\text{dump}$, $\phi_\text{thresh}^{\text{pitch}}$ [deg] &  20, 40 \\
                                    & $R_\text{success}$, $W_\text{target}$ [kg], $\phi_\text{target}$ [deg] &  40, 30, 40 \\
                                    & $\rho^{W}_\text{thresh}$ [kg], $\rho^{\phi}_\text{thresh}$ [deg]& 1, 5 \\

\midrule

\multirow{2}{*}{Observations}       & $d_{\text{end}}^{\text{max}}$ [m] &  0.87  \\
                                    & $x^\text{init}_\text{obs}$, $x_\text{obs}, y_\text{obs}, z_\text{obs}$ [m]  &  0.1, 1.1, 0.4, 0.4 \\

\midrule

\multirow{4}{*}{Control}                 & Control frequency [Hz] & 10 \\

& $[v_\text{wheels}^{\text{min}}, v_\text{wheels}^{\text{max}}]$ [rad/s] & [0, 2.7]\\
    & $[\Dot{\phi}_\text{boom}^{\text{min}}, \Dot{\phi}_\text{boom}^{\text{max}}]$ [rad/s]& [$-$0.191, 0] \\
    & $[\Dot{\phi}_\text{bucket}^{\text{min}}, \Dot{\phi}_\text{bucket}^{\text{max}}]$ [rad/s] & [0, 0.2]\\

\midrule

\multirow{3}{*}{\begin{tabular}{@{}l}
     Muck pile \\
     voxelization 
\end{tabular}}                             & \# of voxels in the $x$-axis, $n_x$ & 60        \\
                                           & \# of voxels in the $y$-axis, $n_y$ & 7         \\
                                           & \# of voxels in the $z$-axis, $n_z$ & 10        \\
                                           
\midrule
\multirow{3}{*}{End zone}                    & $m_\text{min}$ [kg] & 65         \\
                                             & $m_\text{max}$ [kg] & 170        \\
                                             & Minimum height $z_\text{min}$ [m] & 0.35       \\
                                             \midrule
\multirow{4}{*}{\begin{tabular}{@{}l}
     Episodic  \\
     settings 
\end{tabular}}                              & $h_\text{pile}$, $w_\text{pile}$ [m] & 0.68, 0.4\\
   &$\left[\alpha_\text{min}, \alpha_\text{max}\right]$ [deg] & [15, 30] \\
   &$d_\text{attack}$  [m]                     &     [0.1, 0.2]     \\
   &$d_\text{init}$ [m]                    & 0.1 \\

\bottomrule
\end{tabular*}
\label{tab:system-params}
\end{table}

To evaluate the stability of the training procedure, multiple policies with continuous and hybrid action spaces (that is, RLC and RLD policies) are trained using different random seeds. In total, six different seeds are used: five of them are used to train the RL policies (an RLC and an RLD policy for each of these five seeds), and the other seed is used for conducting the evaluation of all the trained models. Table~\ref{tab:system-params} shows the parameters used for the DDPG algorithm, the reward function, the muck pile simulation, and to control general aspects of the problem formulation, such as the episodic settings.

\begin{figure}
    \centering
    \includegraphics[width=\linewidth]{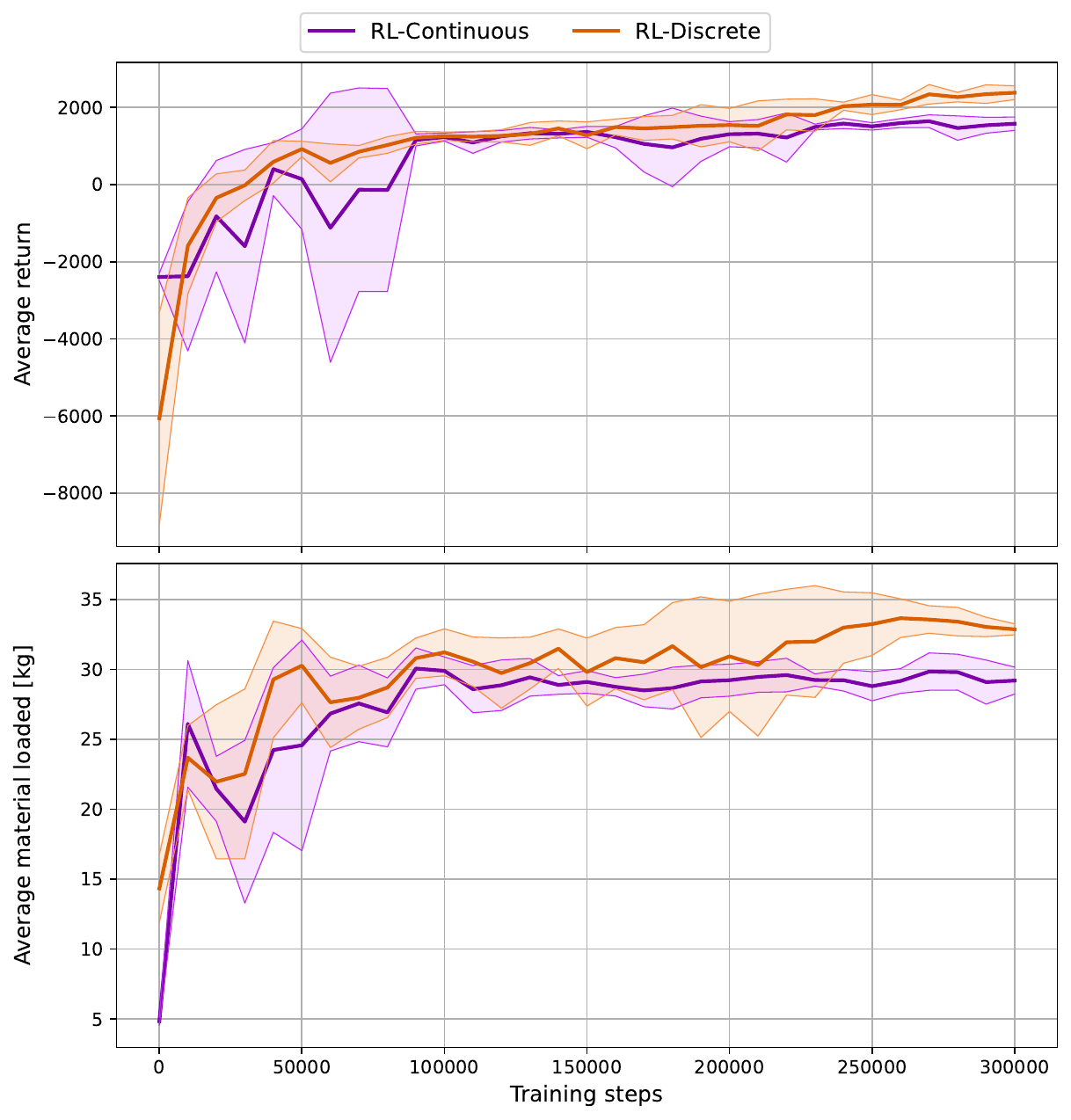}
    \caption{Performance evolution of the trained policies, evaluated in simulation. The curves are constructed averaging the metrics obtained by five independently trained policies per policy type (that is, five RLD and five RLC policies). The shaded areas correspond to the standard deviation of a given metric across the training trials. Note that the AFF metric is not illustrated due to it being derived directly from AML.}
    \label{fig:perf_evolution_sim} 
\end{figure}
\begin{table}
\centering
\caption{Performance achieved by the RLC and the RLD policies by the end of the training process.}
\begin{tabular*}{\linewidth}{l@{\extracolsep{\fill}}ccc}
\toprule
\textbf{Policy} & \textbf{AR} & \textbf{AML [kg]} & \textbf{AFF} \\ \midrule
RLC        & 1,579$\pm$173              & 29.2$\pm$0.9  &   0.97$\pm$0.03               \\
RLD        & 2,384$\pm$178              & 32.9$\pm$0.4  &   1.09$\pm$0.01              \\ \bottomrule 
\end{tabular*}

\label{tab:sim_final_results}
\end{table}

Fig.~\ref{fig:perf_evolution_sim} shows the performance evolution of the RLC and RLD policies during the training process. The curves in this figure are constructed by considering five independent training trials per type of policy, and averaging their performance metrics. As stated previously, said metrics are obtained by evaluating each learned policy for 120 episodes every 10,000 training steps. Table~\ref{tab:sim_final_results}, on the other hand, summarizes the obtained performance metrics for the policies after they are fully trained (i.e., after 300,000 training steps have passed). To account for a fine-grained evolution of performance in terms of compliance with the criteria to consider an episode as successful, refer to Appendix~\ref{appendix:success_level}.

The results displayed in Fig.~\ref{fig:perf_evolution_sim} and in Table~\ref{tab:sim_final_results} show that both RLC and RLD policies are able successfully execute autonomous loading attempts in simulations after they are trained. Fig.~\ref{fig:perf_evolution_sim} shows that, although both types of policies present a similar performance evolution, the RLD policies achieve a better performance by the end of the training process. Table~\ref{tab:sim_final_results} shows that RLD policies outperform RLC policies by 50.9\% in the average return they get, and by 12.7\% in the average amount of material loaded.

\subsection{Real environment evaluation}
\label{sec:real_environment_eval}

For the addressed problem, performing a real world validation allows verifying that the implemented solution (i.e., the reward function design, the simulated environment, and the overall problem formulation) overcomes the reality-gap, that is, that the learned policies can perform successful loading attempts in the real world although they are trained in simulation.

\begin{figure}
\centering
\includegraphics[width=0.9\linewidth]{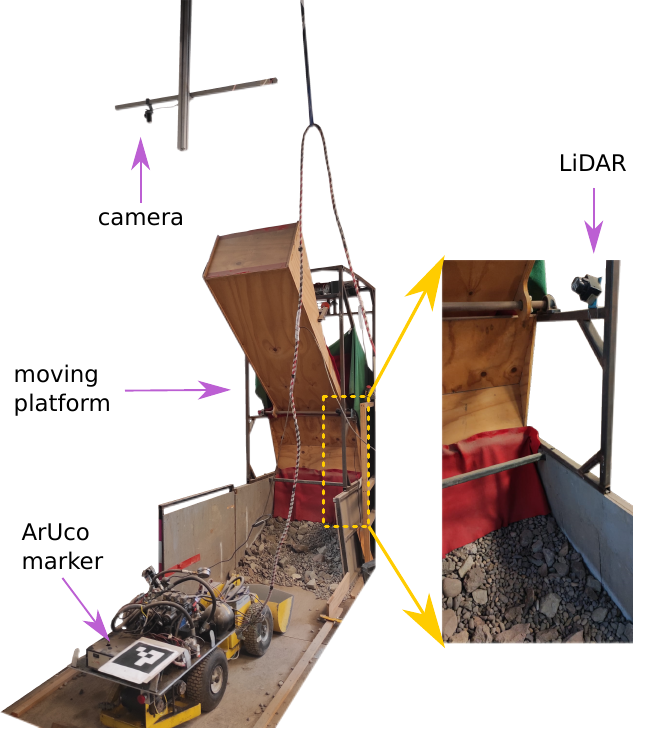}
\caption{Extraction point constructed to conduct real-world experiments.}
\label{fig:scaled_muck_pile_real}
\end{figure}

Fig.~\ref{fig:scaled_muck_pile_real} shows the scaled-down draw point constructed to perform experiments in the real world. This draw point has an actuated moving platform that allows the material to recirculate into the extraction point from above after each loading attempt, and has walls containing the material at floor level so as to simulate a tunnel. Note that this structure is similar to that of a draw point in a block/panel caving mine (see Fig.~\ref{fig:tampier_muck_pile}). 

An ArUco marker~\citep{garrido2016generation} located at the rear of the LHD is used to implement a simple localization system. To do so, a camera mounted over the scaled draw point measures the pose of the ArUco marker at all times. Since the central articulation of the scaled LHD is not actuated during the loading maneuver, variations in the marker's pose are enough to estimate the machine's velocity in the $x$-axis, $v_x$. The marker's pose, alongside the joints' encoders measurements are utilized to estimate the bucket tip pose, which is necessary to compute the distance between the tip of the bucket and the depth limit of the muck pile, $d_\text{end}$. Moreover, a LiDAR sensor mounted over the muck pile is used to estimate its slope, $\alpha$.

For the drift observation, $\mathbb{I}_\text{drift}$, the current used by the electric motors of the wheels during loading is compared to preset values measured in loading attempts without drift, but also by comparing the machine movement in the $x$-axis with the rotation of the wheels, to see if there is a mismatch. If any of these two conditions are true (a power consumption mismatch or if there is wheel rotation but no movement), then $\mathbb{I}_\text{drift}$ is set to 1.

For the real-world validation, the best policies obtained after the training process conducted in simulation are utilized. The action filter used for the loading attempts with the RLD policy is the same used for training, i.e., $\hat{a}^t_\text{boom}$ and $\hat{a}^t_\text{bucket}$ are computed using the filter defined by Eq.~\eqref{eq:action_filter} prior to being denormalized and send as commands to the scaled LHD. However, an action filter is also applied when deploying the RLC policy, which, in this case, limits the numerical range of $a^t_\text{boom}$ and $a^t_\text{bucket}$. To accomplish the above, $a^t_\text{boom}$ is processed such that, if it is greater than 0.6, is directly mapped to 1, and keeps its original value otherwise, whilst $a^t_\text{bucket}$ is compared to $-$0.9, such that values below it are directly mapped to $-$1, and remain unaltered otherwise. This results in no movement in the boom joint whenever $a^t_\text{boom}>0.6$, and no movement in the bucket joint whenever $a^t_\text{bucket}<-0.9$, since, numerically, when $a^t_\text{boom}$ equals 1, its denormalized value corresponds to $\Dot{\phi}^{\text{max}}_\text{boom} =0$, and when $a^t_\text{bucket}$ equals $-$1, its denormalized value corresponds to $\Dot{\phi}^{\text{min}}_\text{bucket} =0$. These filters are used to protect the hydraulic valves against rapid changes in commands that could be sent by the RL controller, and also because the ternary actuators of the LHD would only remain still for fully saturated actions.

\begin{figure}
\centering
\includegraphics[width=0.7\linewidth]{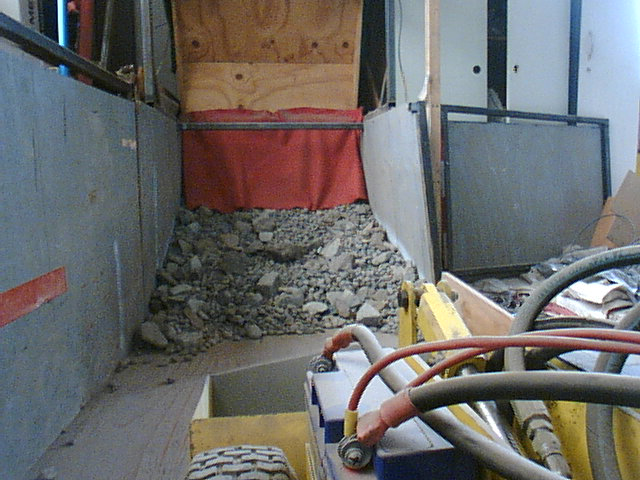}
\caption{Example of the operator's view when performing a teleoperated loading attempt.}
\label{fig:teleop_view}
\end{figure}

The RLC and RLD policies are compared to two other loading systems: a system based on the algorithm developed by~\cite{tampier2021autonomous} (hereafter referred to as the ``\textit{Tampier system}''), and machine teleoperation (hereafter referred to as the ``\textit{Teleop system}''). These two systems are considered as baselines to evaluate the performance of the RL-based policies. For the teleoperation case, we aim at emulating the remote teleoperation of real LHDs, where human operators have mainly visual information, provided by cameras mounted on the LHD, to characterize the environment. Thus, an RGB camera is mounted on the front chassis of the scaled LHD, providing a frontal view as illustrated in Fig.~\ref{fig:teleop_view}. 

The teleoperated loading attempts are performed by three human operators, which possess expert knowledge about strategies for loading material with LHDs. These operators control the LHD using an Xbox controller. 

To measure the performance of the loading systems in the real world, four metrics are utilized:

\begin{itemize}
    \item \textit{Material loaded}: Corresponds to the mass (in kg) of the loaded material after a successful loading attempt. To obtain this metric, after each trial all the material contained in the bucket is measured using a weight scale.
    \item \textit{Elapsed time}: Correspond to the time elapsed during a loading attempt, in seconds.
    \item \textit{Frontal drift}: Corresponds to the percentage of time, with respect to the elapsed time for a given loading trial, in which the frontal wheels of the LHD were drifting.
    \item \textit{Rear drift}: Analogous to the frontal drift metric, but for the rear wheels of the LHD.
\end{itemize}
 
\subsubsection{Loading homogeneous and nonhomogeneous material}
\label{subsubsec:exps_slopes_and_materials}

\begin{figure}
\centering
\subfloat[Homogeneous material]{\includegraphics[width=0.48\linewidth]{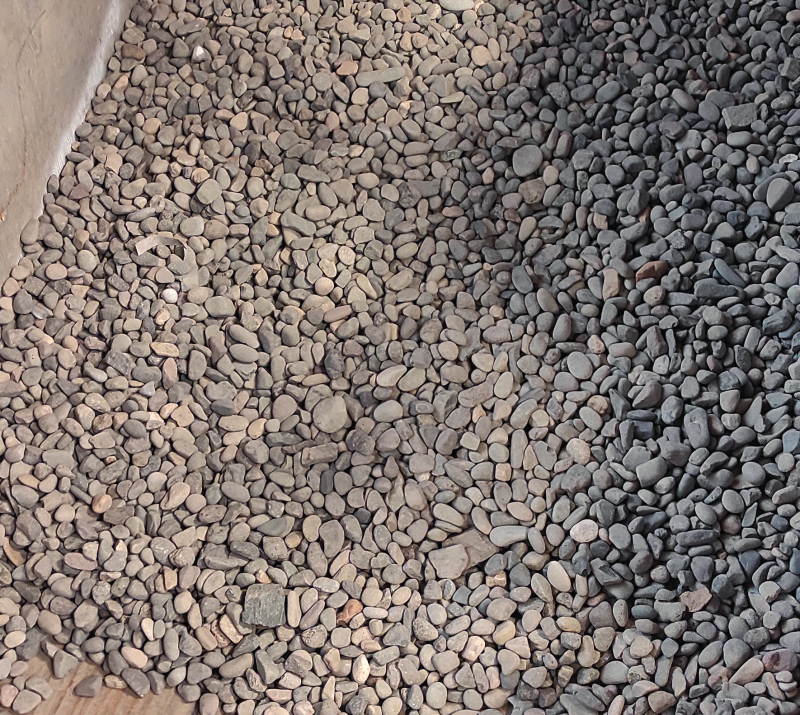}} \quad
\subfloat[Nonhomogeneus material]{\includegraphics[width=0.48\linewidth]{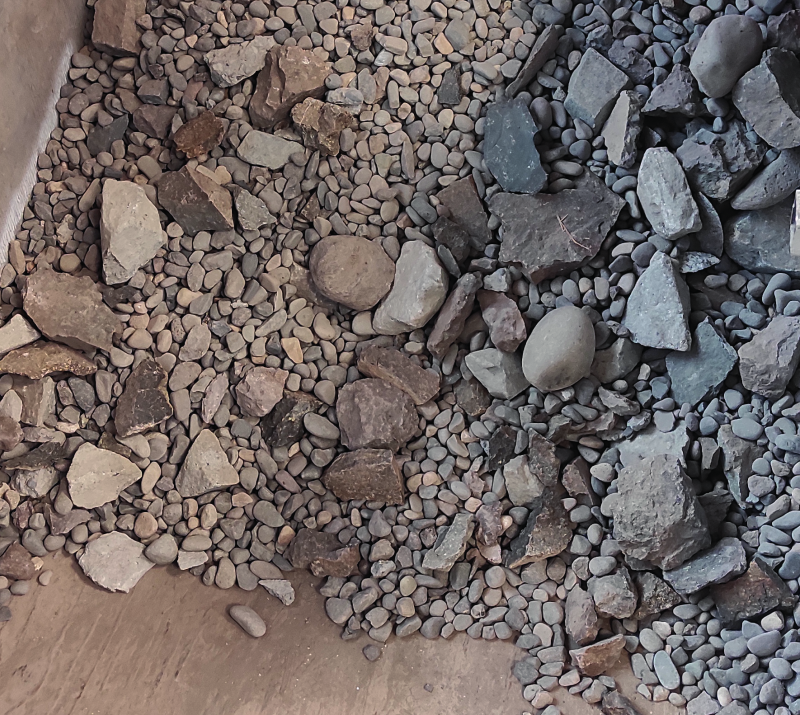}}
\caption{Materials used for the autonomous loading experiments. The homogeneous material consists of gravel with particle sizes varying between 2 and 5~cm. The nonhomogeneous material consists of the gravel from homogeneous material, but with the addition of rocks with sizes between 8 and 15~cm.}
\label{fig:material_comparison}
\end{figure}

To perform loading attempts, both homogeneous and nonhomogeneous material are used. The homogeneous material consists of gravel with particle sizes between 2~cm and 5~cm. The nonhomogeneous material consists of the homogeneous material with the addition of rocks of sizes between 8~cm and 15~cm. Both materials are shown in Fig.~\ref{fig:material_comparison}. To evaluate the performance of the system when loading material from piles with different slopes, the experiments are conducted using piles set to have $\alpha$ value estimations (via LiDAR measurements) approximately equal to 15$^{\circ}$, 20$^{\circ}$, 25$^{\circ}$ and 30$^{\circ}$. 

\begin{table*}
\centering
\caption{Results for the loading attempts performed by all the controllers, in muck piles with different slopes and with homogeneous and nonhomogeneous material.}
\begin{tabular*}{\linewidth}{@{\ }l@{\!\!\!\!\!\!}c@{\ }S[table-format=2.2\pm2.2, parse-numbers=false]@{\extracolsep{\fill}}S[table-format=2.2\pm2.2, parse-numbers=false]@{\extracolsep{\fill}}S[table-format=2.2\pm2.2, parse-numbers=false]@{\extracolsep{\fill}}S[table-format=2.2\pm2.2, parse-numbers=false]@{\extracolsep{\fill}}S[table-format=2.2\pm2.2, parse-numbers=false]@{\extracolsep{\fill}}S[table-format=2.2\pm2.2, parse-numbers=false]@{\extracolsep{\fill}}S[table-format=2.2\pm2.2, parse-numbers=false]@{\extracolsep{\fill}}S[table-format=2.2\pm2.2, parse-numbers=false]@{\ \ }}
\toprule 
& \multirow{2}{*}{\begin{tabular}{c}\textbf{Slope $\boldsymbol{\alpha}$}  \\ \textbf{[deg]} \end{tabular}}& \multicolumn{4}{c}{\textbf{Homogeneous material}} & \multicolumn{4}{c}{\textbf{Nonhomogeneous material}} \\
\cmidrule{3-6} \cmidrule{7-10}
                &        & {\textbf{RLC}}     & {\textbf{RLD}}    & {\textbf{Tampier}}        & {\textbf{Teleop}} & {\textbf{RLC}}     & {\textbf{RLD}}    & {\textbf{Tampier}}        & {\textbf{Teleop}}       \\ 
\midrule
\multirow{4}{*}{\begin{tabular}{@{}l}
     Success  \\
     rate 
\end{tabular}}                        & 15              & {$1.0$}              &   {$1.0$}       & {$0.6$}             & {$1.0$} &  {$1.0$}   &   {$1.0$}      & {$1.0$}    & {$1.0$} \\
                                      & 20              & {$1.0$}              &   {$1.0$}       & {$1.0$}             & {$1.0$} &  {$1.0$}   &   {$1.0$}      & {$0.8$}    & {$1.0$} \\
                                      & 25              & {$1.0$}              &   {$1.0$}       & {$0.8$}             & {$1.0$} &  {$0.8$}   &   {$1.0$}      & {$0.8$}    & {$1.0$} \\
                                      & 30              & {$1.0$}              &   {$1.0$}       & {$0.8$}             & {$1.0$} &  {$0.8$}   &   {$1.0$}      & {$0.8$}    & {$1.0$} \\
                                
\midrule
\midrule
\multirow{5}{*}{\begin{tabular}{@{}l}
     Material  \\
     loaded [kg] 
\end{tabular}}                        & 15              & 22.40\pm2.36   & 23.28\pm1.43   & 23.13\pm2.24  & 25.21\pm0.86 & 24.54\pm0.63   & 25.69\pm0.79   & 25.54\pm1.43  & 25.36\pm1.64 \\
                                      & 20              & 24.18\pm1.22   & 23.48\pm0.95   & 23.87\pm2.41  & 24.27\pm0.73 & 23.99\pm1.15   & 23.74\pm0.77   & 21.34\pm0.80  & 24.36\pm1.15 \\
                                      & 25              & 26.08\pm0.39   & 25.56\pm0.90   & 25.14\pm0.88  & 24.04\pm1.89 & 21.92\pm1.84   & 22.53\pm1.07   & 22.28\pm0.97  & 22.22\pm0.98 \\
                                      & 30              & 27.24\pm0.79   & 28.12\pm0.91   & 25.92\pm0.33  & 25.04\pm1.33 & 25.61\pm2.48   & 24.01\pm2.77   & 23.22\pm0.76  & 25.39\pm1.38 \\
                                      & $\Bar{x}^{a}$   & 24.98\pm1.84   & 25.11\pm1.95   & 24.52\pm1.08  & 24.64\pm0.50 & 24.02\pm1.34   & 23.99\pm1.13   & 23.09\pm1.56  & 24.33\pm1.29 \\
\midrule
\multirow{5}{*}{\begin{tabular}{@{}l}
     Elapsed  \\
     time [s] 
\end{tabular}}             & 15              & 4.06\pm0.23    & 4.38\pm0.50    & 2.58\pm1.35   & 10.42\pm7.03 & 4.10\pm0.50    & 5.21\pm0.66     & 3.16\pm0.31   & 7.23\pm2.44  \\
                                      & 20              & 6.12\pm2.76    & 3.40\pm0.50    & 4.89\pm1.51   & 13.05\pm6.68 & 4.18\pm0.26    & 3.72\pm0.34     & 3.72\pm0.81   & 6.93\pm2.41  \\
                                      & 25              & 3.78\pm0.28    & 4.52\pm0.19    & 5.74\pm0.24   & 16.39\pm8.03 & 3.84\pm0.25    & 3.27\pm0.47     & 3.59\pm0.15   & 8.88\pm2.16  \\
                                      & 30              & 3.98\pm0.20    & 4.72\pm0.36    & 5.56\pm0.10   & 15.91\pm6.84 & 5.30\pm1.10    & 4.68\pm0.64     & 3.97\pm0.90   & 7.69\pm1.18  \\
                                      & $\Bar{x}$       & 4.49\pm0.95    & 4.26\pm0.51    & 4.69\pm1.26   & 13.94\pm2.40 & 4.36\pm0.56    & 4.22\pm0.77     & 3.61\pm0.29   & 7.68\pm0.74  \\
\midrule
\multirow{5}{*}{\begin{tabular}{@{}l}
     Frontal  \\
     drift [\%]
\end{tabular}}                        & 15              & 47.99\pm11.32              & 26.45\pm\hphantom{0}5.75   & 92.53\pm\hphantom{0}9.13  & 28.64\pm10.44             & 39.29\pm\hphantom{0}8.84   & 18.32\pm\hphantom{0}7.69   & 73.11\pm14.39              & 14.47\pm\hphantom{0}9.87  \\
                                      & 20              & 70.59\pm18.82              & 22.94\pm11.38              & 61.67\pm26.55             & 38.10\pm18.08             & 50.31\pm17.18              & 23.30\pm\hphantom{0}6.35   & 77.71\pm13.74              & 13.27\pm\hphantom{0}8.21  \\
                                      & 25              & 25.09\pm\hphantom{0}7.93   & 17.84\pm\hphantom{0}2.92   & 46.19\pm\hphantom{0}3.19  & 36.17\pm18.34             & 59.55\pm27.72              & 21.96\pm11.65              & 75.80\pm13.50              & 22.53\pm18.17 \\
                                      & 30              & 42.62\pm16.33              & 16.32\pm\hphantom{0}3.84   & 36.44\pm\hphantom{0}2.14  & 25.07\pm13.47             & 36.74\pm14.01              & 12.77\pm\hphantom{0}4.54   & 58.86\pm15.33              & 18.20\pm\hphantom{0}8.38  \\
                                      & $\Bar{x}$       & 46.57\pm16.25              & 20.89\pm\hphantom{0}4.04   & 59.21\pm21.24             & 32.00\pm\hphantom{0}5.34  & 46.47\pm\hphantom{0}9.11   & 19.09\pm\hphantom{0}4.08   & 71.37\pm\hphantom{0}7.41   & 17.12\pm\hphantom{0}3.62  \\
\midrule
\multirow{5}{*}{\begin{tabular}{@{}l}
     Rear \\
     drift [\%] 
\end{tabular}}                        & 15              & 35.12\pm19.59              & 52.59\pm\hphantom{0}4.55   & 31.75\pm33.45             & 78.03\pm16.75             & 47.56\pm14.41              & 53.79\pm12.70              & 64.77\pm39.30              & 52.68\pm12.95              \\
                                      & 20              & 88.91\pm\hphantom{0}8.40   & 60.61\pm\hphantom{0}4.01   & 95.77\pm\hphantom{0}4.09  & 75.36\pm21.73             & 74.07\pm\hphantom{0}8.51   & 61.44\pm15.39              & 92.10\pm\hphantom{0}4.91   & 64.40\pm11.64              \\
                                      & 25              & 74.87\pm\hphantom{0}5.46   & 67.35\pm\hphantom{0}4.50   & 96.94\pm\hphantom{0}2.21  & 90.91\pm\hphantom{0}9.76  & 89.93\pm\hphantom{0}4.18   & 61.96\pm\hphantom{0}7.43   & 91.13\pm\hphantom{0}3.55   & 65.64\pm\hphantom{0}5.12   \\
                                      & 30              & 89.24\pm\hphantom{0}6.60   & 88.50\pm\hphantom{0}6.10   & 93.34\pm\hphantom{0}1.64  & 73.56\pm21.06             & 94.13\pm\hphantom{0}6.18   & 80.94\pm\hphantom{0}4.81   & 91.39\pm\hphantom{0}5.99   & 59.34\pm16.13              \\
                                      & $\Bar{x}$       & 72.04\pm22.09              & 67.26\pm13.33              & 79.45\pm27.57             & 79.47\pm\hphantom{0}6.80  & 76.42\pm18.27              & 64.53\pm10.01              & 84.85\pm11.60              & 60.52\pm\hphantom{0}5.10   \\
\bottomrule
\multicolumn{10}{l}{\footnotesize{$^a$Each row $\Bar{x}$ contains the mean value of a given metric across muck piles with different slopes.}} \\
\end{tabular*}
\label{tab:results_summary}
\end{table*}

For all real-world experiments, the initial pose of the LHD is manually set. The center joint is adjusted so that the machine is straight in front of the pile, and the arm is placed in the attack pose (see Section~\ref{subsubsec:episodic_settings}). The tip of the bucket is positioned at 50$\pm$10~cm from the start of the pile, a distance that allows the machine to reach its maximum speed before burying the bucket. For all experiments, the attack speed corresponds to the maximum speed the machine can reach, which is approximately 0.37~m/s. 

Each loading attempt is considered to be successful if the material loaded exceeds 2/3 of the bucket capacity, which in this case corresponds to 20~kg, and the final pose of the bucket is at an angle that allows to pull back from the muck pile without dropping material, i.e., $\phi^T_{\text{shovel}} > \phi^{\text{pitch}}_\text{thresh}$, with $\phi^{\text{pitch}}_\text{thresh}={40}^\circ$. These are the same conditions used for loading attempts in simulation.\footnote{Note that the definition of success used for the real-world experiments is consistent with the success level metric measured in simulation, as explained in Appendix~\ref{appendix:success_level}, for which fulfilling the conditions on loaded material and pitch angle for $\phi_\text{shovel}$ gives an score $\text{SL}\geq 1$.} To protect the electric motors of the scaled down LHD, a timeout of 10~seconds is added to the loading attempts to avoid situations where the machine just tries to move at full power against the muck pile for a long time without performing any actions that move the bucket, which could eventually burn the motors. Loading attempts that end with a timeout and meet both success conditions are considered successes.

For all the real-world experiments, five loading attempts per slope ($\alpha\in\{$15$^\circ$$,$ 20$^\circ$$,$ 25$^\circ$$,$ 30$^\circ$$\}$) and type of material (homogeneous and nonhomogeneous) are performed for the RLC policy, the RLD policy and the Tampier system. For the Teleop system, on the other hand, each operator performs three loading attempts per slope and type of material.\footnote{Example videos of loading trials can be seen in \url{https://youtu.be/jOpA1rkwhDY}.}

Table \ref{tab:results_summary} shows the success rate, average amount of loaded material, average elapsed time per trial, and average measurements of drift (on the frontal and rear wheels) for the performed experiments. The performance metrics for loaded material, execution time, and those associated with drift, are computed considering only successful loading attempts. Note that except for the Tampier system, the RL-based controllers and the Teleop system managed to execute all the loading attempts successfully (getting a success rate equal to 1) for muck piles with homogeneous material. For nonhomogeneous material, only the RLD policy and the Teleop system were able to load successfully in all the trials, while both the RLC policy and the Tampier system failed at some of them. The unsuccessful loading attempts of the Tampier system occurred mainly due to it failing at detecting the initial collision with the pile of material, which resulted in a timeout for some trials since the bucket movement sequence did not trigger (therefore, these loading attempts ended with the scaled LHD not having loaded any material). The failures of the RLC policy in piles with 25$^\circ$ and 30$^\circ$ slopes, on the other hand, were due to timeouts where the machine got stuck in the muck pile without lifting the bucket, and the final material loaded was less than 20~kg.

\begin{figure*}
\centering
\subfloat[Homogeneous material]{\includegraphics[width=0.45\linewidth]{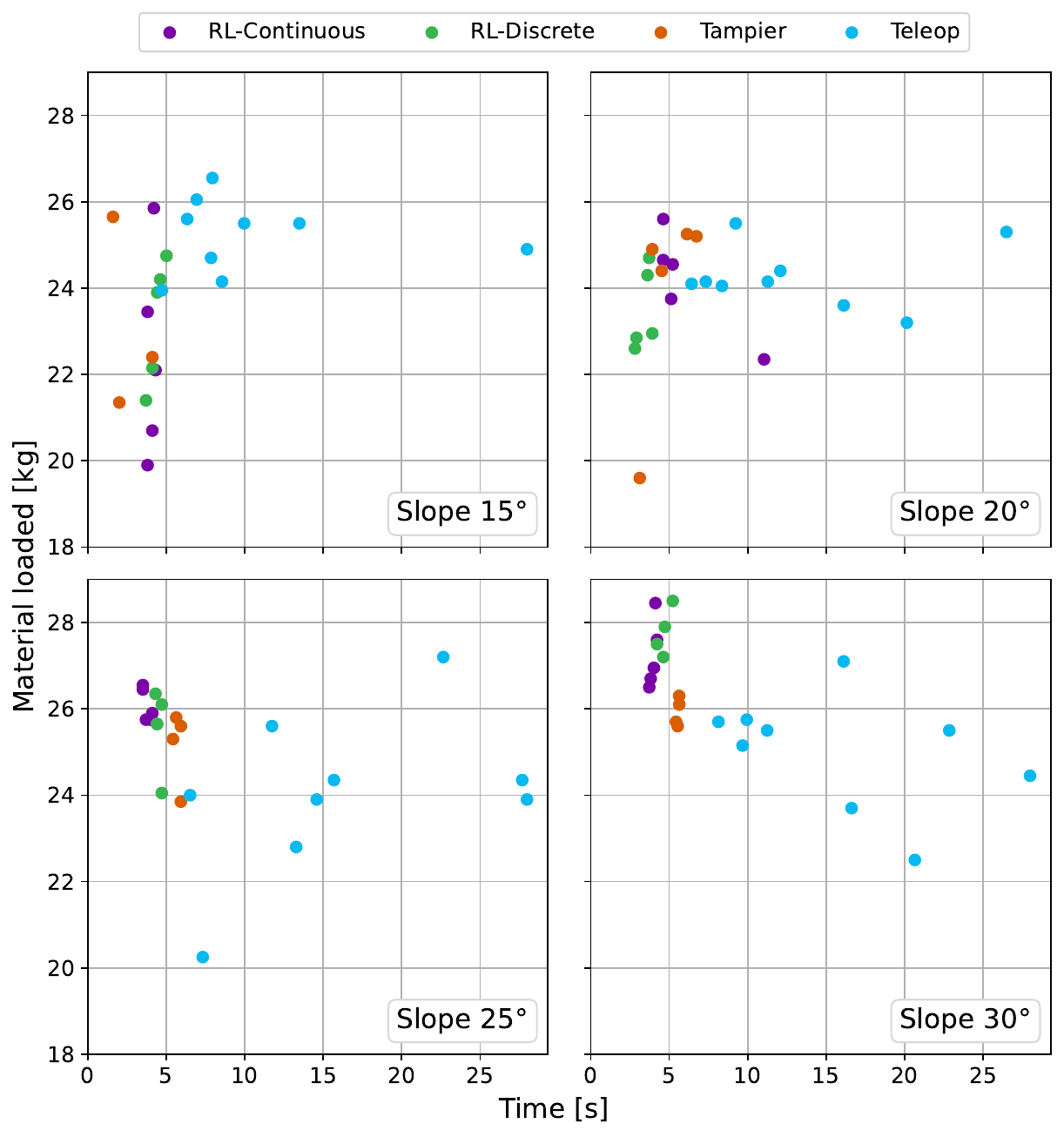}\label{subfig:homogeneous_scatter}}\hfill
\subfloat[Nonhomogeneous material]{\includegraphics[width=0.45\linewidth]{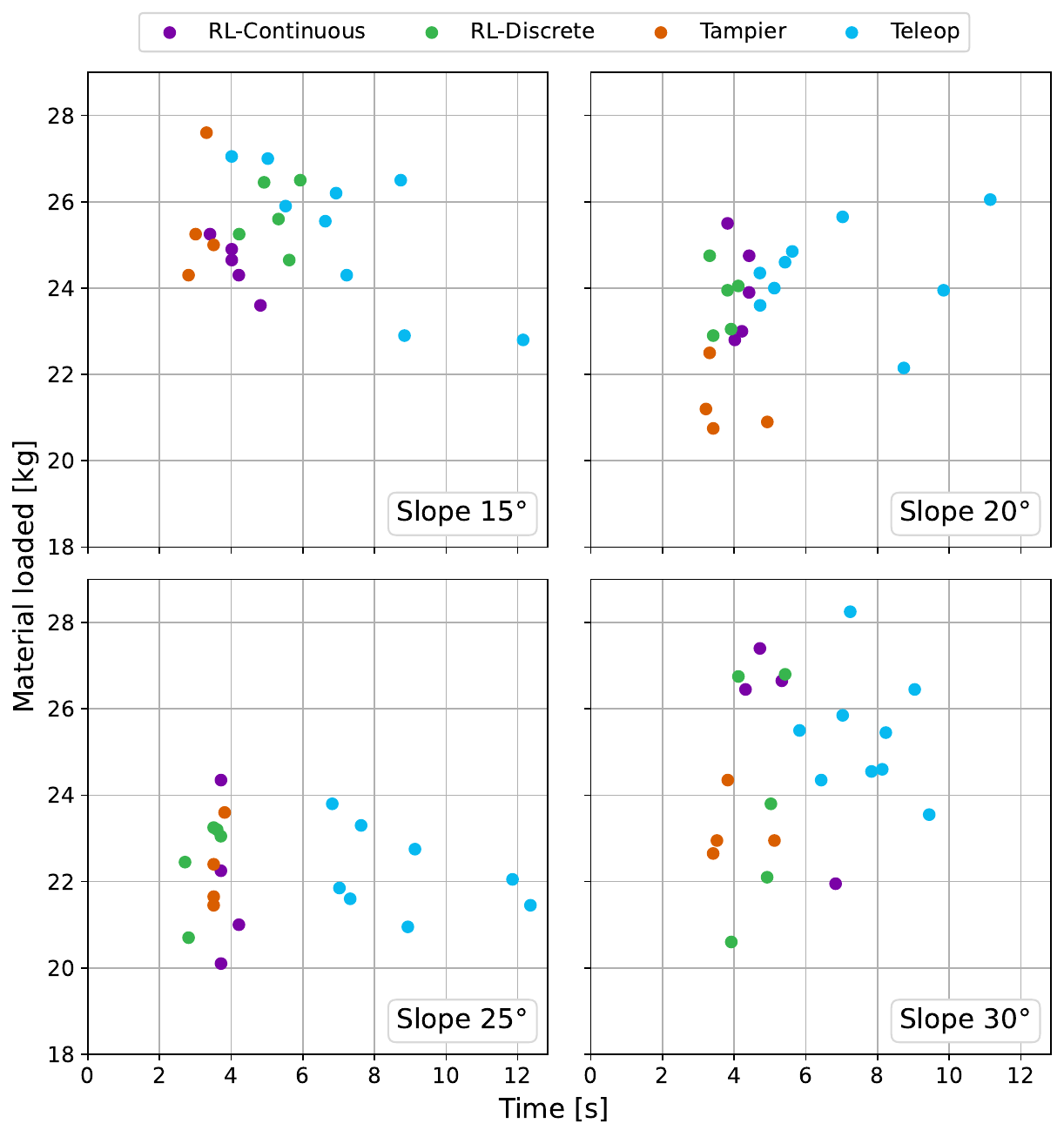}\label{subfig:nonhomogeneous_scatter}}
\caption{Scatter plots for the successful loading attempts performed by all the controllers (the RLC and RLD policies, and the Tampier and Teleop systems), performed in scaled muck piles with (a) homogeneous and (b) nonhomogeneous material.}
\label{fig:loading_attemps_scatter_plot}
\end{figure*}
\begin{figure*}[h!]
\centering
\subfloat[Homogeneous material]{\includegraphics[width=0.45\linewidth]{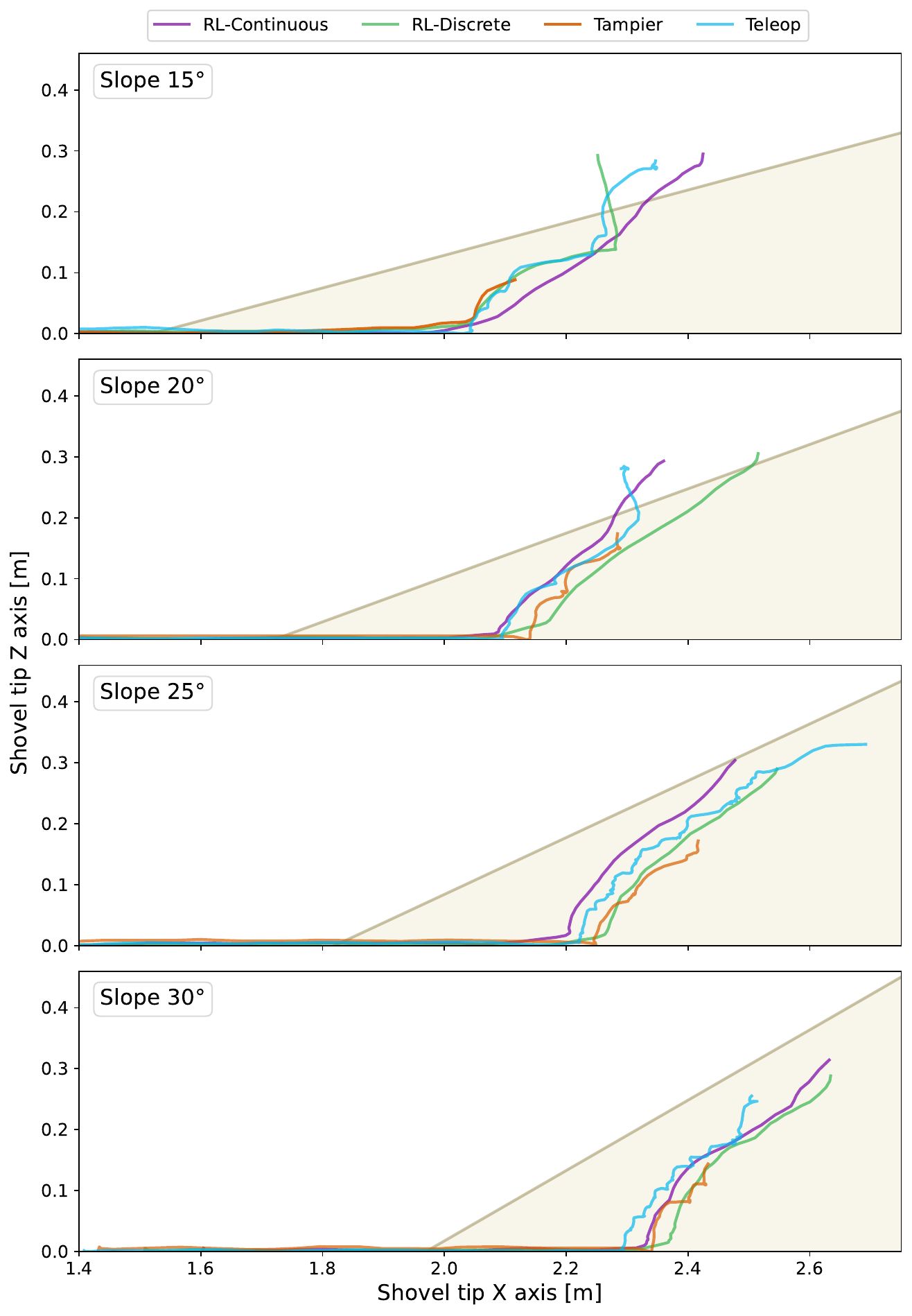}\label{subfig:homogeneous_trajs}} \hfill
\subfloat[Nonhomogeneous material]{\includegraphics[width=0.45\linewidth]{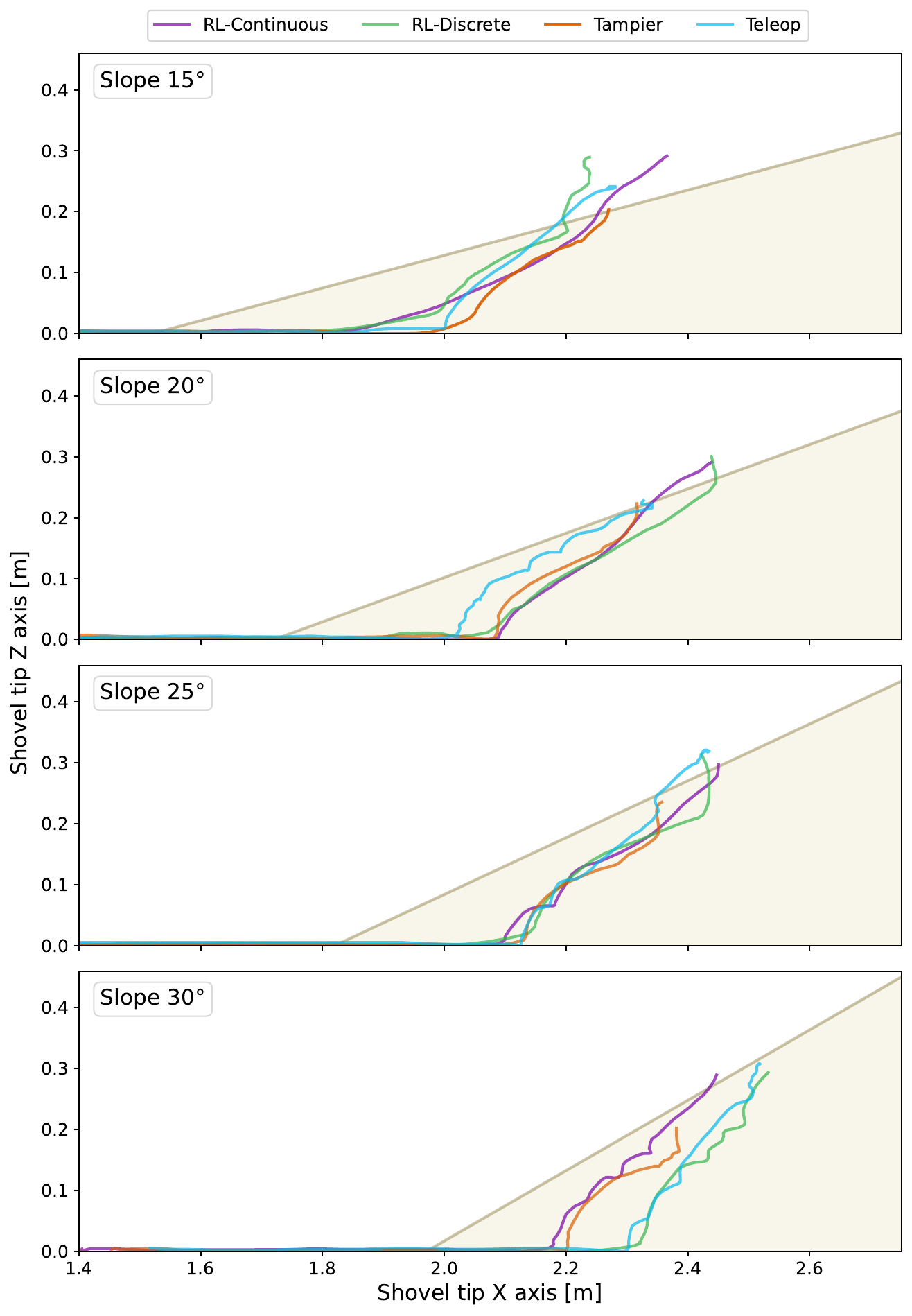}\label{subfig:nonhomogeneous_trajs}} 
\caption{Trajectories of the best loading attempts for each agent and each slope, performed in (a) homogeneous material and (b) nonhomogeneous material.}
\label{fig:best_trajectories}
\end{figure*}

As shown in Table \ref{tab:results_summary}, either loading homogeneous or nonhomogeneous material, the RLD policy obtains the best results overall in the four performance metrics evaluated, and is followed by the RLC policy. 

Regarding the loaded material, all controllers reach values around 25~kg, which corresponds to 83.3\% of the bucket capacity (30~kg). Also note that when loading nonhomogeneous material, the Teleop system achieves the best results in this metric.

For the loading duration, the Tampier system is the fastest, and the Teleop system the slowest, sometimes taking up to more than three times longer than the other controllers per loading attempt. 

As explained in Section~\ref{sec:introduction}, wheel drift should be avoided to minimize the damage to the tires, thus decreasing operational costs. The obtained results show that, regarding the front axle's wheel drift, the RLD policy and the Teleop system achieve the best results, with less dispersion than the RLC policy and the Tampier system. This indicates that the RLD controller and the Teleop system have a better control over when to lift the bucket to regain traction on the front axle when drift is detected. In regards to the rear axle's wheel drift, the four controllers present similar values, however, it is worth noting that the Tampier system has a drift percentage of more than 90\% for slopes of 20$^\circ$, 25$^\circ$ and 30$^\circ$. This may be due to the strategy of the algorithm, which generally never stops accelerating until the end of the excavation process, thus making drifting in the rear wheels likely. 

Fig.~\ref{fig:loading_attemps_scatter_plot} shows scatter plots for all the successful loading attempts performed on the homogeneous and nonhomogeneous muck piles (Fig.~\ref{subfig:homogeneous_scatter} and ~\ref{subfig:nonhomogeneous_scatter}, respectively), in terms of how much material was loaded, and the time each maneuver took.  Fig.~\ref{fig:best_trajectories}, on the other hand, shows the trajectories followed by the tip of the bucket in the XZ plane during the best executed loading attempts for the four controllers, evaluated for different slopes and types of material (homogeneous in Fig.~\ref{subfig:homogeneous_trajs} and nonhomogeneous in Fig.~\ref{subfig:nonhomogeneous_trajs}).

Fig.~\ref{fig:loading_attemps_scatter_plot} shows that the RLC policy, the RLD policy and the Tampier system present similar elapsed times for the loading attempts, as measurements remain near each other in the time axis. It can also be observed that the Teleop system, in contrast, takes the longest amount of time to execute the loading maneuvers, and shows the highest dispersion in this metric. There is also a noticeable difference between the elapsed times associated to teleoperated loading trials in homogeneous and nonhomogeneous material, as the former trials in some cases double the duration of the latter. This can be attributed to experiments being performed first in homogeneous material and then in nonhomogeneous material, allowing the teleoperators to improve their control over the scaled LHD to perform the loading maneuvers better, thus, reducing the time required to successfully complete a trial. 

Fig.~\ref{fig:best_trajectories} provides insights regarding the behaviors learned by the RLC and RLD policies, and how these compare to the heuristic rules in Tampier's loading algorithm and to human expertise. It is observed that the curves followed by the bucket tip for both the RLC and RLD policies are smoother than those produced by the Tampier and the Teleop systems. In fact, the curves followed by the bucket tip in the case of the Tampier system showcase that this method works by following sequential stages: the LHD first buries the bucket into the muck pile, then lifts it and keeps burying it, repeating this pattern until the loading meaneuver is deemed as complete (this is more evident in muck piles with $\alpha$ equal to 20$^\circ$ and 30$^\circ$ in Fig.~\ref{subfig:homogeneous_trajs}). 

From the results presented in this section, it is concluded that the RLD policy, in general terms, outperforms the other baselines, as it is robust to changes to the muck pile slope and the type of material loaded (homogeneous and nonhomogeneous), and also displays a behavior that results in less drift and a low average loading time. The above highlights the importance of conducting the training process taking into account the real-machine dynamics, specially regarding the response of its actuators (since the RLD policies, in contrast to the RLC policies, have a hybrid action space).

\subsubsection{Loading attempts with errors in the slope observation $\alpha$}

\begin{table*}
\centering
\caption{Results for the loading attempts performed on homogeneous material muck piles with a slope of 30$^\circ$ and 15$^\circ$, and errors in the slope observation received by the RLC and the RLD policies.}
\begin{tabular*}{\linewidth}
{@{\ }l@{\extracolsep{\fill}}S[table-format=+2, parse-numbers=false]@{\extracolsep{\fill}}S[table-format=2.2\pm2.2, parse-numbers=false]@{\extracolsep{\fill}}S[table-format=2.2\pm2.2, parse-numbers=false]@{\extracolsep{\fill}}S[table-format=+2, retain-explicit-plus, parse-numbers=false]@{\extracolsep{\fill}}S[table-format=2.2\pm2.2, parse-numbers=false]@{\extracolsep{\fill}}S[table-format=2.2\pm2.2, parse-numbers=false]@{\ \ }}
\toprule
& \multicolumn{3}{c}{\textbf{Muck pile with a slope $\boldsymbol{\alpha}$ of 30$\boldsymbol{^\circ}$}} & \multicolumn{3}{c}{\textbf{Muck pile with a slope $\boldsymbol{\alpha}$ of 15$\boldsymbol{^\circ}$}} \\
\cmidrule{2-4} \cmidrule{5-7}
                     &   {\textbf{Error [deg]}}    & {\textbf{RLC}}      & {\textbf{RLD}}    &  {\textbf{Error [deg]}}    & {\textbf{RLC}}      & {\textbf{RLD}}       \\ 
\midrule
\multirow{4}{*}{Material loaded [kg]}  &   -15              & 23.86\pm0.70     & 25.94\pm0.80    & +15               & 20.76\pm2.66  & 20.59\pm0.85      \\
                                       &   -10              & 27.41\pm1.22     & 25.61\pm1.61    & +10               & 23.13\pm1.23  & 21.77\pm0.45      \\
                                       &   -\hphantom{0}5   & 26.48\pm1.38     & 22.16\pm5.42    & +\hphantom{0}5    & 23.77\pm1.09  & 23.95\pm0.82      \\
                                       &    0               & 27.24\pm0.79     & 28.12\pm0.91    &  0                & 22.40\pm2.36  & 23.28\pm1.43      \\
\midrule    
\multirow{4}{*}{Elapsed time [s]}              &   -15              & 9.71\pm2.22      & 9.93\pm0.04     &  +15              & 3.07\pm0.30   & 4.38\pm0.66       \\
                                       &   -10              & 4.62\pm1.25      & 9.90\pm0.05     &  +10              & 3.56\pm0.22   & 4.02\pm0.29       \\
                                       &   -\hphantom{0}5   & 7.35\pm6.56      & 8.28\pm2.39     &  +\hphantom{0}5   & 3.78\pm0.27   & 4.10\pm0.26       \\
                                       &    0               & 3.98\pm0.20      & 4.72\pm0.36     &  0                & 4.06\pm0.23   & 4.38\pm0.50       \\
\midrule
\multirow{4}{*}{Frontal drift [\%]}      &   -15              & 98.19\pm\hphantom{0}1.36     & 8.42\pm\hphantom{0}1.54     &  +15              & 13.66\pm\hphantom{0}6.95  & 41.63\pm17.34                \\
                                       &   -10              & 41.66\pm14.15                & 26.99\pm33.28               &  +10              & 15.90\pm11.70             & 19.08\pm\hphantom{0}2.90     \\
                                       &   -\hphantom{0}5   & 62.32\pm28.70                & 31.07\pm34.75               &   +\hphantom{0}5  & 40.25\pm18.14             & 7.82\pm\hphantom{0}7.40      \\
                                       &    0               & 42.62\pm16.33     & 16.32\pm\hphantom{0}3.84    &    0              & 47.99\pm11.32             & 26.45\pm\hphantom{0}5.75     \\
\midrule             
\multirow{4}{*}{Rear drift [\%]}       &   -15              & 97.18\pm\hphantom{0}1.05     & 96.60\pm\hphantom{0}1.67    &  +15              & 17.73\pm16.87             & 41.63\pm17.34                \\
                                       &   -10              & 94.59\pm\hphantom{0}1.86     & 95.99\pm\hphantom{0}1.22    &  +10              & 32.02\pm\hphantom{0}9.62  & 24.76\pm\hphantom{0}9.19     \\
                                       &   -\hphantom{0}5   & 94.96\pm\hphantom{0}2.59     & 92.33\pm\hphantom{0}6.87    &  +\hphantom{0}5   & 45.75\pm\hphantom{0}4.94  & 38.34\pm\hphantom{0}3.02     \\
                                       &    0               & 89.24\pm\hphantom{0}6.60     & 88.50\pm\hphantom{0}6.10    &   0               & 35.12\pm19.59             & 52.59\pm\hphantom{0}4.55     \\
\bottomrule
\end{tabular*}
\label{tab:varying_slope_30_15_real}
\end{table*}

The RL-based policies require two observations that provide exteroceptive information about the environment: the slope of the pile, $\alpha$, and the distance between the tip of the bucket and the depth limit, $d_\text{end}$. The $d_\text{end}$ observation component is directly related to the $\alpha$ observation, since the end zone is constructed considering the slope of the muck pile. Thus, it is important that the learned policies are robust to errors in $\alpha$ when loading material. To evaluate how robust the learned policies are to these observation errors, different loading attempts are performed where the slope observation contains artificial measurement errors. To simulate these errors, muck piles with a slope of 15 and 30$^\circ$ are utilized, but the slope observation the agent receives is artificially modified to an incorrect value. The error in the observation is set to vary between 5$^\circ$ and 15$^\circ$. These experiments are only performed for the RLC and RLD policies (using the the same policies and performance metrics for the experiments of Section~\ref{subsubsec:exps_slopes_and_materials}). 

Table \ref{tab:varying_slope_30_15_real} shows the results of loading attempts performed on muck piles with a slope of 30$^\circ$ and 15$^\circ$, and homogeneous material. The ``Error'' column indicates the difference between the slope observation received by the agent and the actual slope of the muck pile. In both cases the observations received by the agents vary between 15$^\circ$ and 30$^\circ$ (the muck pile with a 30$^\circ$ slope has errors varying between $-$15$^\circ$ and 0$^\circ$, whereas the muck pile with a 15$^\circ$ slope has errors varying between 0 and $+$15$^\circ$). For each error in $\alpha$, five loading attempts are performed by each type of policy (RLC and RLD).\footnote{Note that when there is no error in $\alpha$, the data from the experiments reported in Table~\ref{tab:results_summary} for muck piles with a 30$^\circ$ and 15$^\circ$ slope is shown.}$^{,}$\footnote{Some example videos for these experiments using the RLC policy can be seen in \url{https://youtu.be/jOpA1rkwhDY}.}

Regarding the amount of material loaded, the results in Table~\ref{tab:varying_slope_30_15_real} show that for both policies an extreme error in the slope observation ($\pm$15$^\circ$) tend to hinder their performance, as less material is loaded: in this metric, there is a maximum drop of 12.4\% for the RLC policy and of 11.6\% for the RLD policy. 

Interestingly, the detriment in performance is significantly higher for underestimations of $\alpha$. For the muck pile with a 30$^\circ$ slope, when the observation error in $\alpha$ is maximum (i.e., the actual slope is 30$^\circ$ but the policy receives an observation of 15$^\circ$), most loading attempts end by a timeout, with the machine being stuck and trying to move forward without lifting the bucket. This is shown in Table~\ref{tab:varying_slope_30_15_real}, where the loading times increase to almost the maximum (the timeout for real-world tests is 10~seconds as described in Section~\ref{sec:real_environment_eval}, given the control frequency and the maximum number of steps per episode). Also note that in these cases, for the RLC policy the drift percentage in both axles increases almost to the maximum (100\%), but for the RLD policy, while wheel drift on the rear axle increases, the wheel drift for the front axle decreases. This decrease in wheel drift for the RLD policy is due to the machine getting stuck with the bucket of material full and trying to move forward, which generates enough traction so that the front axle cannot turn but with the motors trying to move forward. It is for these cases that the episodic timeout was enforced in the real loading maneuver, so as to protect the scaled LHD's motors. 

For experiments in which $\alpha$ is overestimated, (that is, using the muck pile with a 15$^\circ$ slope, but modifying $\alpha$ with errors of 5$^\circ$, 10$^\circ$ and 15$^\circ$), the results in Table~\ref{tab:varying_slope_30_15_real} show that for the RLC policy, the loading time, wheel drift on both axles and the amount of material loaded all decreased for an extreme error ($+$15$^\circ$) in the slope observation. For the RLD policy, the results follow the same trend, except for front axle drift, which increased, and for the elapsed loading time, which only varied slightly.

As mentioned previously, it is important that the learned policies are robust to errors in the observations's components constructed using exteroceptive information, since this would allow them to be deployed in underground mines with poor visibility. The results obtained in this section show that the loading maneuver is significantly hindered when the policy underestimates the slope of the muck pile, as an extreme underestimation of $\alpha$ causes all loading attempts to end with a timeout and the drift reaches high values. When the policies overestimate the slope of the pile, on the other hand, all loading attempts are completed successfully, although the total amount of material loaded is reduced. Thus, it is found experimentally that the trained RL-based policies are able to withstand an error in $\alpha$ between $-$5$^\circ$ and $+$15$^\circ$ without substantial performance detriments.

\section{Discussion}
\label{sec:discussion}

The results show that the methodology used to train the policies allows obtaining agents capable of performing successful loading maneuvers in muck piles with varying characteristics (different slopes and types of materials) in the real world. To quantitatively assess the loading attempts and compare the performance of the RL-based policies trained in this work with other systems, we consider the amount of material loaded, the elapsed loading time, and the wheels' drift as key metrics. 

The experiments performed in this work show that the RL-based policies outperform the Tampier controller (a heuristics-based method) in loading both homogeneous and non-homogeneous material. The main improvement observed is with respect to the wheels' drift metric, where the RLC and RLD policies show a low percentual drift per loading attempt for both of the LHD axles. When comparing the RL-based policies to the Teleop system, both outperformed the Teleop system in the homogeneous muck pile, but when loading non-homogeneous material, the Teleop agents obtained slightly better results in all metrics, except for the elapsed loading times. One can attribute the above to the fact that a human operator, having access to richer exteroceptive information, is able to deal more easily and reactively with undesirable situations (such as drifting). Additionally, rich visual information about the muck pile allows operators to plan ahead for a proper loading strategy. 

Regarding the experiments performed to assess the robustness of the policies to errors in the muck pile slope estimation, $\alpha$, the results obtained show that, while both RL-based policies are capable of performing successful loading maneuvers, underestimations in $\alpha$ are more disadvantageous than overestimations. Indeed, underestimations may make the controllers attempt to exit the muck pile too early, which results in the machine getting stuck. Furthermore, although there are only two observations related to the environment, the observations regarding the position of the LHD bucket in the $x$-axis, and the wheel drift indicator, also allow the agent to identify when to start (or stop) lifting the bucket to progress on the maneuver, thus compensating for errors in $\alpha$.

Finally, since the RL-based policies are trained through multiple interactions with the environment (in this case for 300,000 time steps), a high-speed simulation is crucial so that the training process does not take an excessive amount of time. In this work, an extension of the analytical FEE model is used to simulate the interaction between the LHD and the material to be loaded, which allows for a fast simulation. One of the main limitations of using the extended FEE, however, is that it does not use a contact physics model for the interaction between the bucket and the material. The total force applied to the LHD is calculated based solely on the bucket tip pose within the pile. The above results in situations where the bottom of the bucket is in contact with material, but since the bucket tip is outside the muck pile, no force is applied on the bucket, which generates unrealistic situations. These cases mainly affect the training of the RL agents during the initial phase of the training process (where an exploratory behavior is encouraged), and $r^t_\text{bottom}$ is added to the reward function definition so as to penalize the agent if it reaches these scenarios (and to prevent it from exploiting them). A principled solution to this problem would be using a more capable simulator, such as the one utilized in \cite{backman2021continuous}, however, at the cost of potentially increasing the required time for training the policies.

\section{Conclusion}
\label{sec:conclusion}

In this work we propose a method for learning how to load from a muck pile with an LHD by formulating the problem as a POMDP, and solving it using deep reinforcement learning. A simulation of the interaction between the machine and the muck pile is also proposed. The performance of the trained policies is compared with a heuristic-based loading controller and with loading maneuvers performed by human operators via teleoperation. Additional experiments were performed to evaluate the robustness of the policies when there are errors in the pile slope measurements.

The obtained RL-based policies are able to perform loading maneuvers successfully in the real world after being trained in simulation, outperforming the baseline controllers in the majority of scenarios and key metrics. These trained policies also displayed good performance when artificial errors where introduced in the estimation of the muck pile slope they observed. 

Two main directions for future work are identified: (i) improving the simulation and (ii) testing the policies on a full-scale machine. Improving the simulation of the interaction between the material to be loaded and the LHD is a critical point to improve the results, as it would allow a simplification of the reward function (e.g. by removing the term associated to burying the bottom of the bucket in the pile), and could facilitate obtaining better results by further bridging the sim2real gap. A first step to improve the implemented simulation could be to add discrete elements to the FEE formulation utilized; a methodology used in \cite{holz2015advances}. This new implementation of the simulation could also allow for the addition of machine hydraulic pressure observations. Finally, testing the trained policies on a full-scale machine is necessary to truly evaluate the performance of the policies, as these tests will show if they can successfully perform loading maneuvers when deployed in LHD machines used in real mining operations, and if any adjustments need to be made to the problem formulation proposed in this work.

\appendix
\section*{Average ``success level'' during training}
\label{appendix:success_level}

To measure the performance evolution of the trained policies considering the criteria to label an episode as successful, the 
\textit{Average success level} (ASL) metric is utilized. This metric is computed as the average value, across evaluation episodes, of the score defined by Eq.~\eqref{eq:success_rate_levels}, where $\phi_\text{shovel}$ is the pitch angle of the bucket (see Fig.~\ref{fig:lhd_angles_diagram}), $\phi^{\text{pitch}}_\text{thresh}$ is a threshold for said angle, and $W^T$ is the weight of the material loaded at the end of an evaluation episode. Note that $W^T$ is compared to values that are near the bucket capacity in kg (which, for the scaled LHD, is equal to 30 kg).
\begin{equation}
        \text{SL} =  
          \begin{cases} 
           5 & \text{if } (W^T \geq 28) \land  (\phi_\text{shovel}^{T}>\phi^{\text{pitch}}_\text{thresh}), \\
           4 & \text{if } (28 > W^T \geq 26) \land  (\phi_\text{shovel}^{T}>\phi^{\text{pitch}}_\text{thresh}),   \\
           3 & \text{if } (26 > W^T \geq 24) \land  (\phi_\text{shovel}^{T}>\phi^{\text{pitch}}_\text{thresh}),   \\
           2 & \text{if } (24 > W^T \geq 22) \land (\phi_\text{shovel}^{T}>\phi^{\text{pitch}}_\text{thresh}),   \\
           1 & \text{if } (22 > W^T \geq 20) \land  (\phi_\text{shovel}^{T}>\phi^{\text{pitch}}_\text{thresh}),  \\
           0 & \text{otherwise.}
          \end{cases}
    \label{eq:success_rate_levels}
\end{equation}

The motivation behind the definition of ASL, is that to consider a loading attempt as successful, both the material loaded and the final pose of the bucket must be considered. In this regard, the LHD should load as much material as possible, but also has to end the episode with the bucket at an angle that allows it to pull back from the muck pile without potentially dropping material. Note that the position of the bucket is already evaluated by dividing the muck pile into zones in the XZ plane (see Fig.~\ref{fig:muck_pile_zones}).

\begin{figure}
    \centering
    \includegraphics[width=\linewidth]{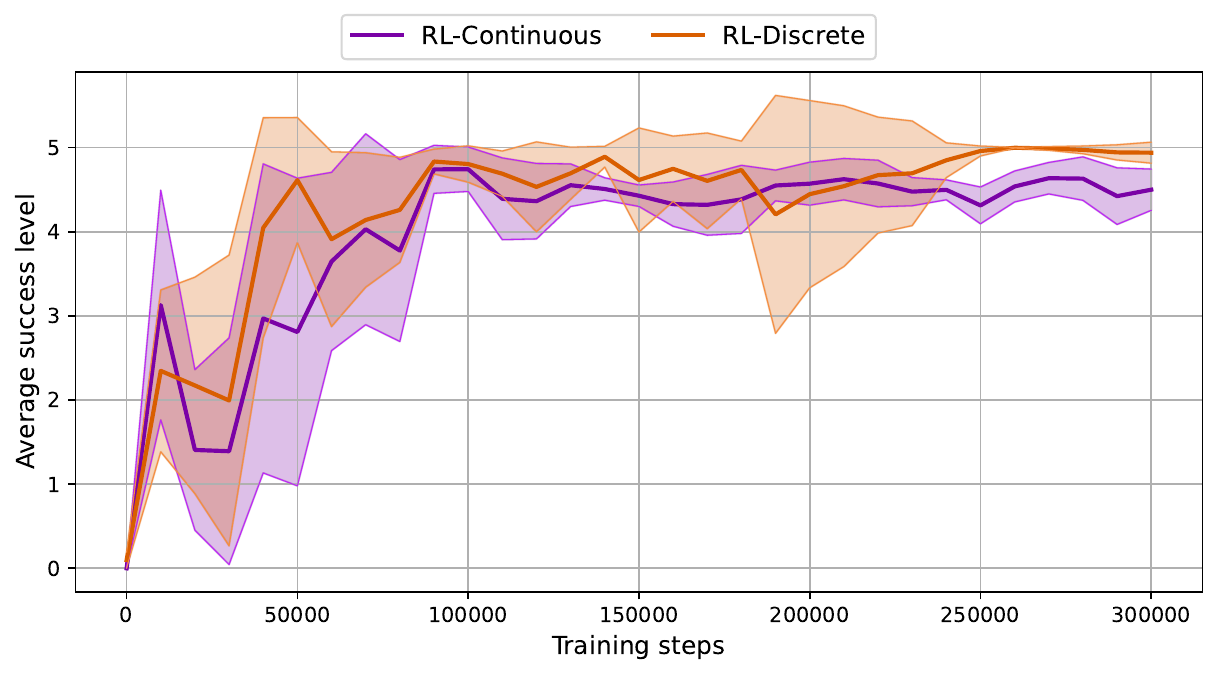}
    \caption{Performance evolution of the trained policies in terms of the ``Average success level'' metric, evaluated in simulation. The curves are constructed as in Fig.~\ref{fig:perf_evolution_sim}. The shaded areas correspond to the standard deviation across the training trials.}
    \label{fig:asl_evolution}
\end{figure}

The evolution of the ASL metric during training is shown in Fig.~\ref{fig:asl_evolution} (five independent training trials for both types of policies, as explained in Section~\ref{subsec:sim_training_and_eval}). By the end of the training process, i.e., after 300,000 steps, the ASL is equal to 4.50$\pm$0.25 for the RLC policies, and equal to 4.95$\pm$0.12 for the RLD policies.

While the RLD policies also outperform the RLC policies in terms of the metrics reported in Section~\ref{subsec:sim_training_and_eval}, the results obtained for the RLD policies in terms of the ASL metric show that they also consistently achieve the targets defined by $r^T_\text{success}$, that is, to get to load the bucket capacity, and to reach a pose for the LHD arm so that the machine can pull back from the muck pile after the loading maneuver is done.

\section*{Acknowledgments}
The authors would like to thank Claudio Palacios and Gonzalo Olguín for their help during the execution of some of the experiments conducted using the scaled-down LHD.

\section*{Declaration of conflicting interests}
The authors declare that they have no known potential conflicts of interest with respect to the research, authorship, and/or publication of this article.

\balance 

\bibliographystyle{apalike}
\bibliography{references}

\end{document}